\ifdefined\pdfminorversion\pdfminorversion=7\fi
\documentclass[10pt,twocolumn,letterpaper]{article}
\usepackage[T1]{fontenc}

\usepackage[pagenumbers]{cvpr}

\usepackage{colortbl}
\definecolor{ourgray}{gray}{0.92}

\usepackage{multirow}
\usepackage{makecell}

\newcommand{\tablestyle}[2]{\setlength{\tabcolsep}{#1}\renewcommand{\arraystretch}{#2}\centering\footnotesize}
\newcommand{\tablestylesm}[2]{\setlength{\tabcolsep}{#1}\renewcommand{\arraystretch}{#2}\centering\scriptsize}
\newcommand{\up}{$\uparrow$}
\newcommand{\dn}{$\downarrow$}
\newcommand{\sd}[1]{\underline{#1}}

\newcommand{\grp}[2]{\multicolumn{#1}{l}{\textbf{#2}}}

\newcommand{\ours}{\rowcolor{ourgray}}
\newcommand{\x}{$\times$}

\usepackage{xspace}
\newcommand{\method}{CodecOPSD\xspace}
\newcommand{\numcond}{16\xspace}
\usepackage{stfloats}
\usepackage{placeins}

\definecolor{cvprblue}{rgb}{0.21,0.49,0.74}
\usepackage{fontawesome5}
\definecolor{projaccent}{RGB}{176,38,82}
\usepackage[pagebackref,breaklinks,colorlinks,allcolors=cvprblue]{hyperref}

\def\confName{CVPR}
\def\confYear{2027}

\newcommand{\projecturl}{https://kiteretsu77.github.io/CodecOPSD_website/}
\newcommand{\projectlink}{{\hypersetup{urlcolor=projaccent}\href{\projecturl}{\textcolor{projaccent}{\faGlobe}\hspace{0.5em}\textsf{\textbf{\textcolor{projaccent}{Project Page}}}\hspace{0.7em}\textsf{\textcolor{black!65}{kiteretsu77.github.io/CodecOPSD\_website}}}}}

\title{\LARGE On-Policy Codec Distillation\\ for Efficient All-in-One Video Generation\vspace{-12pt}}

\author{%
Boyang Wang\textsuperscript{1,2*}\hspace{1.3em}Kevin Blackburn-Matzen\textsuperscript{1}\hspace{1.3em}Oindrila Saha\textsuperscript{1}\\[2pt]
Matheus Gadelha\textsuperscript{1\textdagger}\hspace{1.3em}Zezhou Cheng\textsuperscript{2\textdagger}\\[4pt]
{\small \textsuperscript{1}Adobe Research\qquad \textsuperscript{2}University of Virginia}\\[6pt]
{\normalsize\projectlink}
}

\begin{document}
\twocolumn[{%
\renewcommand\twocolumn[1][]{#1}%
\vspace*{-0.5in}
\maketitle
\vspace{-9mm}
\begin{center}
\captionsetup{type=figure}
\includegraphics[width=\textwidth]{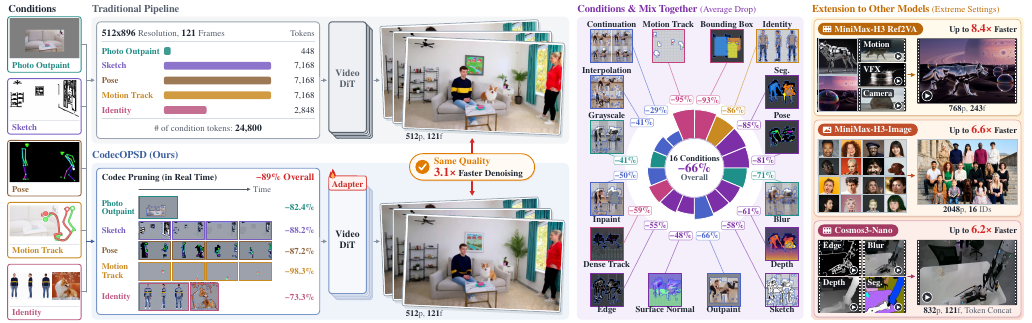}

\vspace{4pt}
\captionof{figure}{
\textbf{\method keeps the informative condition tokens and shows a clear speedup.}
Our codec pruning drops 89\% of the condition tokens with consistent quality, while being 3.1\x\ faster.
Our method simply adds an adapter to the video DiT through our proposed on-policy codec distillation.
The same pruning works on \numcond condition types and is extendable to MiniMax-H3 Ref2VA~\cite{minimax2026h3}, \mbox{MiniMax-H3-Image}~\cite{minimax2026h3image}, and Cosmos3~\cite{agarwal2026cosmos}, with up to 8.4\x\ faster denoising on higher resolution, longer frames, and more conditions.
}

\label{fig:teaser}
\end{center}
\vspace{3mm}
}]
{\renewcommand{\thefootnote}{\fnsymbol{footnote}}\footnotetext[1]{Work partially done during an internship at Adobe Research.\\\hbox to 1.8em{\hss\textsuperscript{\textdagger}}Equal advising.}}

\begin{abstract}
Modern visual generative models operate on tokens, and every visual asset they receive, whether an identity, a scribble, a depth map, or a motion video, is tokenized and appended to the input token sequence.
Consequently, conditioning quickly dominates the computation cost.
Yet most of these tokens carry redundant spatiotemporal information.
Our key insight is that identifying what is genuinely new is exactly the problem that classical video codecs have long solved.
Thus, we first introduce a codec-based token pruning method that discards redundant tokens.
Then, we propose an on-policy self-distillation (OPSD) strategy to train a DiT adapter capable of learning the codec pruning pattern well without quality degradation.
Together, they form On-Policy Codec Distillation (\method).
We scale \method to \numcond condition modalities on our own video diffusion model, where it prunes 66\% of the condition tokens on average with almost no quality drop.
Therefore, denoising becomes 2\x\ faster, and the cache memory shrinks by up to 73\% in KV cache mode.
\method also extends to other state-of-the-art pretrained image and video generative models, where its benefit grows further with more conditions, longer frames, and higher resolution.
\end{abstract}

\section{Introduction}
\label{sec:intro}

Creativity rarely arrives in text alone.
To turn an afternoon at home into a video, a creator brings a photo of the living room sofa, a sketch of the room around it, the poses of two people, a picture of the family corgi, and its path scribbled across the floor (Fig.~\ref{fig:teaser} left).
State-of-the-art video generative models can receive all these
conditions as input.
However, modern generative models concatenate all condition tokens into one sequence~\cite{ben2026avcontrol,ju2025fulldit,ju2026editverse} as inputs, so the attention cost grows quadratically with every added condition.
Fine-grained controllability then leads to severe computational cost.

Most existing acceleration methods speed up the denoiser through fewer sampling steps~\cite{yin2024one}, FlashAttention~\cite{Shah2024FlashAttention3FA}, quantization~\cite{Micikevicius2022FP8FF}, feature caching~\cite{Liu2024TimestepET}, or sparse attention~\cite{Xi2025SparseVA}.
We instead reduce the condition tokens at their source, which is orthogonal to and pluggable into these methods.
Condition signals are highly compressible, since a static background repeats in every frame and a motion path covers a few pixels.
In our daily life, every video we watch or store is compressed by a video codec such as H.264~\cite{wiegand2003overview}, which removes exactly this redundancy.
These codecs are the product of decades of traditional video compression research: rule-based systems that run efficiently on nearly every phone, camera, and browser.
For every pixel of every frame, a codec records how it can be predicted from earlier frames, as a motion vector, and what remains new, as a residual.
These signals thus provide a natural rule for selecting condition tokens of every modality, from natural RGB images and videos to sketches, depth, and motion tracks.
They can be computed in real time on a CPU during training and inference.

Turning this idea into a working model takes two steps: selecting the right tokens and adapting the model to learn from them.
First, raw codec signals cannot be used directly and need careful engineering.
We resolve these issues, propose an algorithm to map the signals from pixels to VAE tokens, and adapt the selection to each modality.
Second, pruning alone is not enough, as the model must learn the codec pattern for faithful generation.
We observe that supervised fine-tuning (SFT) leaves a clear gap to the dense model on codec-pruned conditions.
We instead propose adapting the model to the pruned conditions through on-policy self-distillation (OPSD)~\cite{Zhao2026SelfDistilledRO}, which supervises the model on the states of its own rollouts rather than on noised ground truth~\cite{Agarwal2023OnPolicyDO}.
Teacher and student are the same base model, but the teacher additionally sees the tokens dropped by codec pruning as inferable privileged information.
The teacher turns these tokens into dense guidance at every state of the student's own rollout.
We name this method On-Policy Codec Distillation (\method).

To demonstrate the effectiveness of \method{}, we adapt it to \numcond condition modalities in our video diffusion model, namely segmentation, bounding box, identity, sketch, edge, depth, surface normal, human pose, sparse and dense tracking, blurring, grayscale, first-last frame interpolation, continuation, inpainting, and outpainting.
With quality close to the dense teacher, it drops 66\% of the condition tokens on average, which speeds up denoising by up to 2.30\x\ and shrinks up to 73\% of the memory cost in KV cache mode.
We further apply it to three state-of-the-art models, namely MiniMax-H3 Reference-to-Audio-Video (Ref2VA)~\cite{minimax2026h3} for video generation, MiniMax-H3-Image~\cite{minimax2026h3image} for image generation, and Cosmos3~\cite{agarwal2026cosmos} for physical AI tasks.
At similar quality, the denoising speedup keeps growing with the input scale, reaching 3.72\x\ on MiniMax-H3 Ref2VA with more references, 3.06\x\ on MiniMax-H3-Image at $2048^2$, and 2.30\x\ on Cosmos3, with consistent gains across tasks as diverse as control-to-video, video continuation, video-to-audio, and video-to-action generation.

Our contributions are summarized as follows:
\begin{itemize}[leftmargin=*,nosep]
\item We propose codec pruning in controllable generation, which maps the motion vectors and residuals of a standard H.264 codec from pixels to VAE token masks.
\item We introduce \method, where the teacher additionally sees the codec-pruned tokens as privileged information, leaving the student an information gap that is inferable and of the right size.
\item We scale \method to \numcond condition modalities on our video diffusion model, and further extend it to the state-of-the-art all-in-one models MiniMax-H3 Ref2VA, MiniMax-H3-Image, and Cosmos3.
\item We show that \method outperforms other token pruning baselines and has consistent gains in denoising speed with consistent quality.
\end{itemize}

\section{Related Work}
\label{sec:related_work}

\paragraph{Controllable Video Generation.}
Early controllable image and video generators add one condition type at a time, such as ControlNet~\cite{zhang2023adding}, SparseCtrl~\cite{guo2024sparsectrl}, and AVControl~\cite{ben2026avcontrol}.
Later, more works extend them to multiple conditions for richer creative expression~\cite{alhaija2025cosmos,wang2026frame,qi2025over++,zhou2026tri}.
On the other hand, all-in-one models train one network on many condition types. VACE~\cite{jiang2025vace} injects a unified condition module through a context adapter, FullDiT~\cite{ju2025fulldit} and EditVerse~\cite{ju2026editverse} place all inputs in one token sequence under self-attention, and UniVideo~\cite{wei2026univideo} and OmniWeaving~\cite{pan2026omniweaving} pair a multimodal language model with the diffusion transformer.
The recent MiniMax-H3 Ref2VA~\cite{minimax2026h3} accepts several image and video conditions at the same time.
These models mostly append conditions along the sequence rather than the channel dimension, so as users bring ever more references, quadratic attention dominates the cost.
This is exactly the problem we are addressing in this work.

\paragraph{Token-Pruning-based Compression in Vision.}
In diffusion models, ToMe-SD~\cite{bolya2023token} merges tokens inside the network, and FullDiT2~\cite{he2025fulldit2} keeps the condition tokens ranked highest by a learned scoring network.
Keep The Essentials~\cite{parihar2026keep} instead drops reference tokens before the network, at random or by DeepGaze~\cite{kummerer2025modeling} saliency.
Another common practice is to downscale the condition, as in OminiControl2~\cite{tan2026ominicontrol2}, EasyControl~\cite{zhang2025easycontrol}, and AVControl~\cite{ben2026avcontrol}.
In vision-language models, codec-based encoders such as OneVision-Encoder~\cite{tang2026onevision}, Video-LaVIT~\cite{jin2024video}, CoPE-VideoLM~\cite{sarkar2026cope}, CodecSight~\cite{Zou2026CodecSightLV}, and AdaCodec~\cite{Hou2026AdaCodecAP} read the motion vectors and residuals of the compressed stream to skip or compactly encode redundant patches.
These VLM works target understanding, where dropping redundant patches rarely changes the answer, so they can prune aggressively.
Controllable generation is inherently different, since the content of a pruned token still appears in the output, which calls for a more conservative selection.
To the best of our knowledge, we are the first to use codec signals for token pruning in diffusion models, where we map them to 3D VAE tokens and adapt the selection to each condition modality.

\paragraph{On-Policy Distillation.}
In large language models, on-policy distillation~\cite{Agarwal2023OnPolicyDO} trains a student on its self-generated sequences with token-level teacher feedback.
On-policy self-distillation~\cite{Zhao2026SelfDistilledRO} lets one model act as both teacher and student, where the teacher additionally sees privileged information.
In diffusion models, D-OPSD~\cite{Jiang2026DOPSDOS} fine-tunes a step-distilled image model without losing its few-step ability, where the teacher additionally sees the target image.
OPSD-V~\cite{Liu2026OPSDVOS} post-trains few-step autoregressive video models on their own rollouts, where the teacher instead sees real-video context.
DanceOPD~\cite{zhou2026danceopd} distills several teachers into one student, and supervises a single state of each student rollout.
PDM~\cite{Li2026RethinkingCG} and LIFT~\cite{Ji2026LIFTLV} distill dense per-frame control into sparse control.
In our work, the privileged information is the condition tokens dropped by codec pruning, where the model learns to preserve quality from a fraction of the original condition tokens.

\section{Method}
\label{sec:condition_pruning}
\begin{figure}[t]
\centering
\includegraphics[width=\columnwidth]{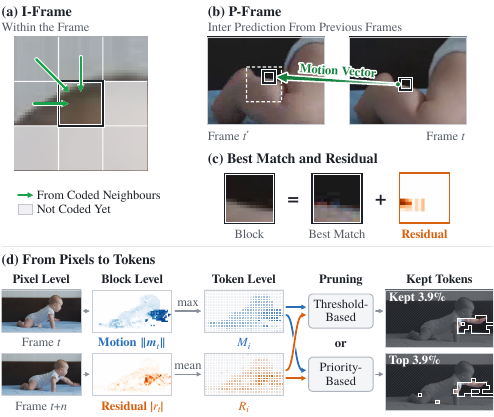}
\vspace{-8pt}

\caption{\textbf{From codec signals to kept tokens.}
(a) An I-frame predicts each block from its coded neighbours.
(b) A P-frame finds the best match of each block in a previous frame and stores its motion vector.
(c) The residual is the difference between the block and its best match.
(d) Each token receives a motion score and a residual score from the blocks it covers, and threshold-based or priority-based pruning then selects the kept tokens.
}
\label{fig:codec}

\end{figure}

\begin{figure*}[t]
\centering
\vspace{-4pt}
\includegraphics[width=\textwidth]{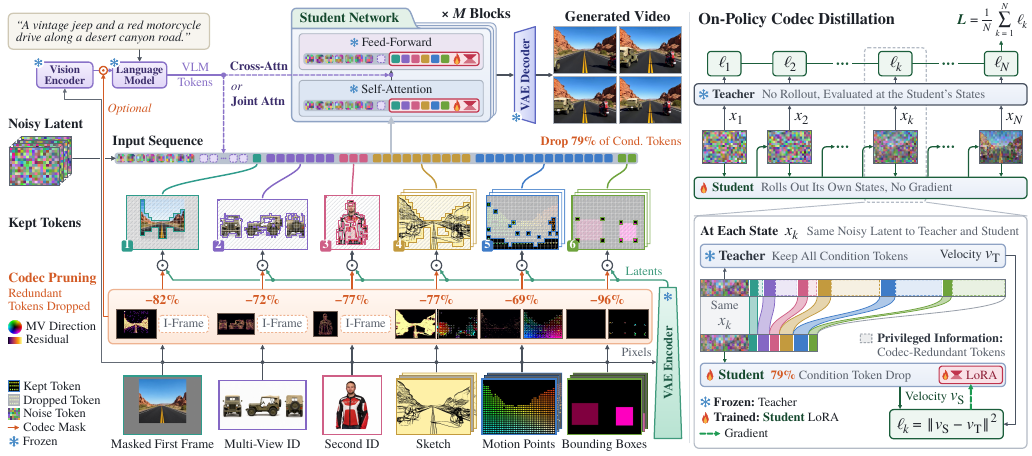}
\vspace{-7pt}

\caption{\textbf{Pipeline of \method.}
{\textbf{Left:} Codec pruning runs in parallel with the VAE encoder and drops condition tokens at a dynamic rate per condition, and the kept tokens concatenate with the noisy latent as the DiT input.
It optionally drops the vision tokens of a VLM, which enter the DiT by cross-attention or joint attention depending on the model.
\textbf{Right:} On-Policy Codec Distillation provides the tokens removed by codec pruning to the frozen teacher as privileged information and compares the student's prediction with the teacher's.}}
\label{fig:pipeline}

\end{figure*}

We first review the video codec preliminaries in Sec.~\ref{sec:prelim} (Fig.~\ref{fig:codec}).
In Sec.~\ref{sec:codec_token_mask}, we show how we turn the codec signals from the pixel to the token level, and in Sec.~\ref{sec:kept_tokens}, we illustrate the methodology of feeding the kept tokens to the DiT model.
In Sec.~\ref{sec:opcd}, we design the model training for \method.
The pipeline is shown in Fig.~\ref{fig:pipeline}.

\subsection{Video Codec Preliminaries}
\label{sec:prelim}

As shown in Fig.~\ref{fig:codec}, a video codec such as H.264~\cite{wiegand2003overview} predicts each block of pixels from a reference block in the same frame or in past frames.
While a codec also involves transform, quantization, and entropy coding, these stages are not relevant to our method, and we focus on how the reference and the residual of each block are obtained.

The codec divides the frames of a video into two types: intra-coded frames (I-frames) and predictive frames (P-frames).
An I-frame has no temporal reference and predicts each block from its neighbours within the same frame (Fig.~\ref{fig:codec}a).
A P-frame predicts each block from previously coded frames, where the compression encoder finds the best match with classical block-matching search algorithms (Fig.~\ref{fig:codec}b).
This match is rarely exact, so the codec stores the residual, which is the difference between the ground-truth block and its best match (Fig.~\ref{fig:codec}c).
In summary, a P-frame $t$ provides a motion vector $m_t(p)$ and a residual $r_t(p)$ for the block of each pixel $p$, while an I-frame provides a residual and no motion vector.

\subsection{Token Pruning from Codec Signals}
\label{sec:codec_token_mask}

In video diffusion models, the 3D VAE compresses the input both spatially and temporally, where a token $c_i$ of the DiT can cover a set of $32\times32$ pixels $\mathcal P_i$ over a set of 8 frames $\mathcal F_i$ in pixel space.
In contrast, the motion vectors $m_t$ and residuals $r_t$ obtained from the codec are at a much finer level, from $16\times16$ down to $4\times4$ pixel blocks.

As shown in Fig.~\ref{fig:codec}d, we first extract the motion vectors $m_t(p)$ and the residuals $r_t(p)$ from the codec signal, and then map them onto the spatiotemporal region of each token $c_i$.
To adapt to different resolutions, we also rescale these signals by the condition size.
To convert dense information into a scalar per token, we compute a motion score $M_i$ and a residual score $R_i$ by:
\begin{equation}
  \begin{aligned}
  M_i&=\max_{t\in\mathcal F_i}\max_{p\in\mathcal P_i}\|m_t(p)\|,\\
  R_i&=\max_{t\in\mathcal F_i}\operatorname*{mean}_{p\in\mathcal P_i}\big|r_t(p)\big|.
  \end{aligned}
  \label{eq:scores}
\end{equation}
Based on these two scalar scores on each token, we propose two pruning mechanisms in this paper: a threshold-based and a priority-based one to adapt to different demands.

First, we propose a threshold-based pruning mechanism because we observe that videos with high camera motion are harder to compress than videos that are almost static.
Therefore, it is suitable to find a balance point for each video through a threshold-based design, which we formulate as:
{\setlength{\abovedisplayskip}{3pt}\setlength{\belowdisplayskip}{3pt}\setlength{\abovedisplayshortskip}{0pt}\setlength{\belowdisplayshortskip}{3pt}%
\begin{equation}
  b_i=
  \begin{cases}
  1, & \text{if } M_i\ge\tau_m \text{ or } R_i>\tau_r,\\
  0, & \text{otherwise},
  \end{cases}
  \label{eq:admissible}
\end{equation}}%
where $\tau_m$ and $\tau_r$ are the motion and residual thresholds, and $b_i=1$ means that we will keep the current token $c_i$.

Second, we also propose a priority-based pruning mode for flexible budget demands.
We construct a sortable energy $E_i$ formula by combining $R_i$ and $M_i$ by:
\begin{equation}
  {E_i=\bar R_i\,\big(1+\lambda \bar M_i\big)},
  \label{eq:gazer}
\end{equation}
where $\bar R_i$ and $\bar M_i$ are normalized $R_i$ and $M_i$ in the range $[0,1]$, and $\lambda$ is a scalar weight for the motion.
In this design, we sort the tokens by $E_i$ value and drop the lowest first.

\subsection{Conditioning with Token Selection}
\label{sec:kept_tokens}

Our conditions span diverse modalities, and to adapt them in one system, each condition $j$ is rendered as an image or a video $I^{(j)}$ and then encoded by the same frozen VAE encoder $\mathcal E$ into condition tokens $\mathcal C^{(j)}=\mathcal E\big(I^{(j)}\big)$, as shown in Fig.~\ref{fig:pipeline} left.
In parallel, codec pruning computes the kept mask $b^{(j)}$ following Sec.~\ref{sec:codec_token_mask}.
The DiT follows standard flow matching~\cite{lipman2022flow} and denoises the noisy latent $x_\sigma=(1-\sigma)z+\sigma\epsilon$ of the generated video, where $z$ is the clean latent, $\epsilon$ is Gaussian noise, and $\sigma$ is the noise level.
The mask is applied to each condition, and the kept tokens of the $J$ conditions are concatenated with $x_\sigma$ as the input sequence $\mathcal S$. The formula is:
\begin{equation}
  \begin{aligned}
  \tilde{\mathcal C}^{(j)}&{}=b^{(j)}\odot\mathcal C^{(j)},\quad j=1,\dots,J,\\
  \mathcal S&{}=\big[\,x_\sigma;\ \tilde{\mathcal C}^{(1)};\dots;\ \tilde{\mathcal C}^{(J)}\,\big],
  \end{aligned}
  \label{eq:sequence}
\end{equation}
where $\odot$ applies the mask element-wise and $x_\sigma$ is not pruned.
Each kept token retains its original 3D RoPE~\cite{su2024roformer} position, and the resulting sequence $\mathcal S$ is fed into the DiT without architectural modification.

Meanwhile, state-of-the-art models such as MiniMax-H3 Ref2VA and MiniMax-H3-Image~\cite{minimax2026h3,minimax2026h3image} also feed the condition references to a VLM encoder (optional path in Fig.~\ref{fig:pipeline}).
Codec pruning reuses the same codec signals to prune these vision tokens too.
The vision encoder (ViT) of the VLM still reads every reference in full, and codec pruning drops its output vision tokens before they enter the language model of the VLM.

\begin{table*}[!t]
\definecolor{t1light}{RGB}{232,240,252}

\caption{
\textbf{Token Pruning Comparisons.}
Under the same average drop rate, we compare different token pruning methods in either training-free or fine-tuned settings.
Our \method shows better numerical performance on most metrics and stays close to the teacher.
Prep.\ is the token selection time in seconds per case, and ours runs on 6 CPU cores of an Intel Xeon Platinum 8275CL.
The best results in each group are \textbf{highlighted}, and the second best are \sd{underlined}.
}

\label{tab:token_selection}
\vspace{-3pt}

\tablestyle{1pt}{1.05}
\newlength{\tonecol}\settowidth{\tonecol}{\scriptsize In/Outpaint}%
\newlength{\tonetmp}\settowidth{\tonetmp}{\scriptsize PCK@0.1\,$\uparrow$}\ifdim\tonetmp>\tonecol\setlength{\tonecol}{\tonetmp}\fi%
\newlength{\tonemeth}\settowidth{\tonemeth}{\footnotesize SFT, Reduced Res.}\addtolength{\tonemeth}{5pt}%
\newlength{\toneside}\setlength{\toneside}{10pt}%
\newlength{\toneprep}\settowidth{\toneprep}{\scriptsize Prep.}%
\settowidth{\tonetmp}{\scriptsize (s)\,$\downarrow$}\ifdim\tonetmp>\toneprep\setlength{\toneprep}{\tonetmp}\fi%
\setlength{\tabcolsep}{\dimexpr(\textwidth-\tonemeth-\toneprep-12\tonecol-\toneside)/30\relax}%
\begin{tabular}{>{\centering\arraybackslash}p{\toneside} >{\raggedright\arraybackslash}p{\tonemeth} >{\centering\arraybackslash}p{\toneprep} *{12}{>{\centering\arraybackslash}p{\tonecol}}}
\toprule
& & & \multicolumn{3}{c}{Quality and Fidelity} & \multicolumn{9}{c}{Condition Following} \\
\cmidrule(lr){4-6}\cmidrule(l){7-15}
& \scriptsize Method & \scriptsize\makecell{Prep.\\(s)\,\dn} & \scriptsize FID\,\dn & \scriptsize FVD\,\dn & \scriptsize LPIPS\,\dn & \scriptsize \makecell{Identity\\DINOv3\,\up} & \scriptsize \makecell{Segm.\\mIoU\,\up} & \scriptsize \makecell{Depth\\si-RMSE\,\dn} & \scriptsize \makecell{Normal\\Angle\,\dn} & \scriptsize \makecell{Edges\\F1\,\up} & \scriptsize \makecell{In/Outpaint\\mPSNR\,\up} & \scriptsize \makecell{Pose\\PCK@0.1\,\up} & \scriptsize \makecell{Track\\TrajErr\,\dn} & \scriptsize \makecell{Grayscale\\$\Delta E_{00}$\,\dn} \\
\midrule
& Teacher & 0 & 22.4 & 118.0 & 0.273 & 0.288 & 0.888 & 0.647 & 4.72 & 0.445 & 20.93 & 0.912 & 13.54 & 6.63 \\
\midrule
& Random & $\sim$0 & 29.2 & 161.2 & \sd{0.351} & 0.278 & 0.782 & \sd{0.959} & \sd{7.44} & 0.286 & 17.88 & 0.870 & 15.73 & \sd{8.65} \\
& Reduced Res. & $\sim$0 & 56.3 & 381.1 & 0.663 & 0.269 & 0.288 & 3.740 & 30.05 & 0.066 & 10.73 & 0.286 & 38.42 & 27.35 \\
& DINO~\cite{Simoni2025DINOv3} & 23.8 & \sd{28.8} & 151.1 & 0.368 & \sd{0.284} & \textbf{0.838} & 1.053 & 8.13 & \sd{0.321} & \sd{18.21} & \sd{0.891} & \textbf{11.69} & 9.59 \\
& Saliency~\cite{parihar2026keep} & 75.0 & 30.0 & \sd{148.5} & 0.401 & \textbf{0.286} & \sd{0.796} & 1.470 & 11.88 & 0.294 & 15.69 & \textbf{0.895} & 13.91 & 13.01 \\
\rowcolor{t1light}\cellcolor{white}\multirow{-5}{*}{\rotatebox[origin=c]{90}{\scriptsize\textit{Training-free}}} & Codec (ours) & 1.13 & \textbf{24.5} & \textbf{132.3} & \textbf{0.307} & \sd{0.284} & 0.734 & \textbf{0.868} & \textbf{6.05} & \textbf{0.393} & \textbf{19.89} & 0.836 & \sd{13.90} & \textbf{6.93} \\
\midrule
& SFT, Random & $\sim$0 & 25.6 & 142.7 & 0.305 & \textbf{0.289} & 0.864 & 0.875 & 6.22 & 0.338 & 19.11 & 0.901 & \sd{7.50} & \sd{6.78} \\
& SFT, Reduced Res. & $\sim$0 & 34.7 & 199.5 & 0.444 & \sd{0.288} & 0.655 & 1.633 & 14.43 & 0.114 & 14.72 & 0.835 & 14.53 & 12.62 \\
& SFT, DINO~\cite{Simoni2025DINOv3} & 23.8 & 26.7 & 138.0 & 0.337 & 0.287 & \textbf{0.885} & 0.941 & 7.45 & 0.343 & 18.62 & \sd{0.910} & \textbf{6.86} & 8.16 \\
& SFT, Saliency~\cite{parihar2026keep} & 75.0 & 29.6 & 146.4 & 0.398 & 0.281 & 0.805 & 1.427 & 11.87 & 0.297 & 15.78 & 0.898 & 13.32 & 12.70 \\
\rowcolor{t1light}\cellcolor{white}& SFT, Codec (ours) & 1.13 & \sd{23.8} & \sd{130.9} & \sd{0.297} & 0.284 & 0.796 & \sd{0.767} & \sd{5.80} & \sd{0.404} & \sd{19.83} & 0.890 & 11.03 & 7.24 \\
\rowcolor{t1light}\cellcolor{white}\multirow{-6}{*}{\rotatebox[origin=c]{90}{\scriptsize\textit{Fine-tuned}}} & \method (ours) & 1.13 & \textbf{23.0} & \textbf{113.8} & \textbf{0.278} & \textbf{0.289} & \sd{0.883} & \textbf{0.706} & \textbf{5.44} & \textbf{0.417} & \textbf{20.43} & \textbf{0.911} & 7.80 & \textbf{6.53} \\
\bottomrule
\end{tabular}
\end{table*}

\subsection{On-Policy Codec Distillation}
\label{sec:opcd}

Our training design is based on on-policy self-distillation (OPSD), and the full pipeline is shown in Fig.~\ref{fig:pipeline} right.
OPSD uses a single base model as both teacher and student, where the teacher supervises the student on the states of the student's own rollout.

The student shares the same base model $v_\theta$ as the teacher, but with an additional trainable LoRA~\cite{hu2021lora} $\phi$.
As shown in Fig.~\ref{fig:pipeline}, the LoRA is applied to the projections of the condition tokens, but not to those of the noisy latent or the VLM tokens.
The student takes the kept tokens $\tilde{\mathcal C}$, where the mask follows the threshold-based pruning of Eq.~\ref{eq:admissible} during training.
The teacher is frozen and receives the same condition types as the student, with no extra condition type.
It additionally takes the dropped tokens $\mathcal C^{\star}=\mathcal C\setminus\tilde{\mathcal C}$ of these conditions, which serve as its privileged information.
Accordingly, their velocity predictions are formulated as:
\begin{equation}
  \begin{aligned}
  v_\mathrm{T}(x,\sigma)&=v_\theta(x,\sigma;\tilde{\mathcal C}\cup\mathcal C^{\star}),\\
  v_\mathrm{S}(x,\sigma)&=v_{\theta,\phi}(x,\sigma;\tilde{\mathcal C}).
  \end{aligned}
  \label{eq:teacher_student}
\end{equation}
The student learns to match the teacher without this privileged information, which is feasible since the kept tokens are the most representative ones of each condition and can largely represent the dropped regions.

In each training step, the student first rolls out its own trajectory without gradient.
Starting from pure noise $x_1=\epsilon$, it denoises step by step along the $K$ decreasing noise levels of the inference sampler:
\begin{equation}
  x_{k+1}=x_k+(\sigma_{k+1}-\sigma_k)\,v_\mathrm{S}(x_k,\sigma_k),
  \label{eq:rollout}
\end{equation}
where $\sigma_k$ is the $k$-th noise level, with $\sigma_1=1$ for pure noise, and $x_k$ denotes the $k$-th state of the rollout.
We supervise a subset $\mathcal N$ of these states, namely the first $N$ states at high noise.
At each supervised state, the teacher and the student receive the same $x_k$ and $\sigma_k$, and we minimize the distance between their predictions:
\begin{equation}
  \begin{aligned}
  \ell_k&={\big\|v_\mathrm{S}(x_k,\sigma_k)-v_\mathrm{T}(x_k,\sigma_k)\big\|_2^2},\\
  \mathcal L_{\mathrm{OPSD}}(\phi)&={\mathbb E_{\mathcal C,\epsilon}\Big[}\frac{1}{|\mathcal N|}\sum_{k\in\mathcal N}\ell_k{\Big]},
  \end{aligned}
  \label{eq:opsd}
\end{equation}
where the expectation is over the training conditions and the noise.
Note that no ground truth enters Eq.~\ref{eq:opsd}, since the supervision comes from the teacher.
At inference, the student runs with $\tilde{\mathcal C}$, and the teacher is no longer needed.

\section{Experiments}
\label{sec:experiments}

Secs.~\ref{sec:exp_setup} and~\ref{sec:exp_eval} provide the implementation and evaluation details.
Sec.~\ref{sec:exp_token_selection} analyzes token dropping and speedup on our multi-condition video model, and Sec.~\ref{sec:exp_gap} provides an ablation.
Sec.~\ref{sec:exp_other_models} extends \method to MiniMax-H3 Ref2VA~\cite{minimax2026h3}, the H3-Image variant~\cite{minimax2026h3image}, and Cosmos3~\cite{agarwal2026cosmos} for multi-task generation.

\subsection{Implementation details}
\label{sec:exp_setup}

We build our own multi-condition video model as the main testbed of this work.
It is relatively lightweight and not distilled, and its training data comes with ground truth, so it can be directly trained with SFT.
This enables the extensive SFT comparisons in Tab.~\ref{tab:token_selection} and shows the versatility of our method across diverse visual controls, each with its own metric.
Our model starts from a pretrained 22B-parameter text-image-to-video DiT, and its condition tokens never attend to the noise tokens, so it supports KV cache~\cite{tan2026ominicontrol2}.
For the first stage, it is initially trained with text and a mix of \numcond condition modalities, namely segmentation, bounding box, identity, sketch, edge, depth, surface normal, human pose, sparse and dense tracking, blurring, grayscale, frame interpolation, continuation, inpainting, and outpainting.
Each training sample uses 1 to 5 of them.
All token pruning methods in Tab.~\ref{tab:token_selection} are also based on these weights.
In the second stage, it is trained with \method for 2K steps on 16 H200 GPUs.
Inference sampling uses 30 denoising steps, and \method rolls out the first 8 of these 30 steps in training.
Further implementation details are in the supplementary material.

\subsection{Evaluation details}
\label{sec:exp_eval}
\begin{figure}[t]
\centering
\includegraphics[width=\columnwidth]{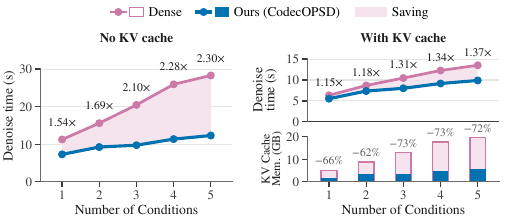}
\vspace{-12pt}

\caption{\textbf{Denoising time and memory change.}
Test cases run at 512\x896 with 121 frames, 8 sampling steps~\cite{yin2024one}, and one H200.
Ours is faster both without the KV cache (left) and with the KV cache (right top).
The KV cache requires extra memory cost, and ours needs much less of it (right bottom).}
\label{fig:stress}

\end{figure}

\begin{figure*}[!b]
\vspace{-1pt}
\centering
\includegraphics[width=\textwidth]{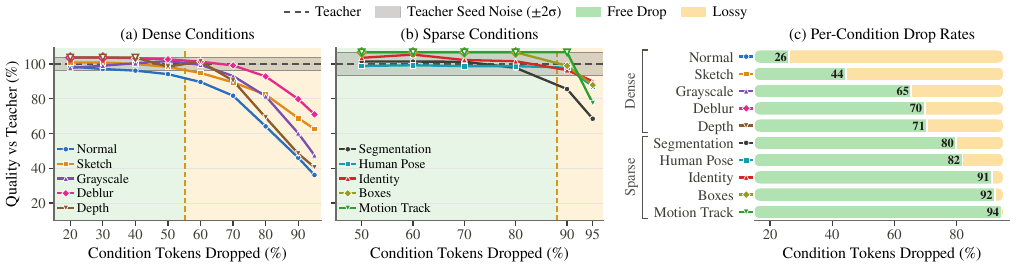}
\vspace{-5.0mm}

\caption{\textbf{Per-condition drop rates.}
We drop the tokens of a single condition at different rates in the order of Eq.~\ref{eq:gazer} and measure quality relative to the teacher in each target metric, for (a) dense conditions and (b) sparse conditions.
The grey band is the teacher's seed noise ($\pm2\sigma$).
The green region marks the free drop, where quality stays within this band, and the orange region marks the lossy drop.
(c) provides the free drop value of each condition.
}
\label{fig:percond}

\end{figure*}

\begin{figure*}[!tp]
\centering
\vspace{-3.0mm}
\includegraphics[width=\textwidth]{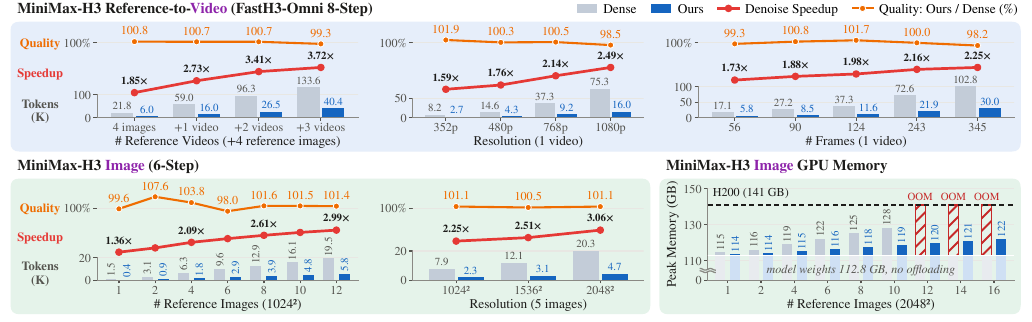}
\vspace{-5.0mm}

\caption{\textbf{Scaling tokens on MiniMax-H3 Ref2VA and MiniMax-H3-Image.}
\textbf{Top:} MiniMax-H3 Ref2VA with more reference videos, higher resolution, and more frames; the speedup keeps growing without a clear quality drop relative to the dense teacher.
For both speedup and quality, MiniMax-H3 Ref2VA runs the 8-step FastH3-Omni of FastVideo~\cite{pmlr-v267-zhang25m}.
\textbf{Bottom:} MiniMax-H3-Image with more reference images and higher resolution, which has the same trend.
At $2048^2$ pixels, with every module (DiT, VLM, and VAE) loaded on the GPU and no offloading, ours still fits 16 images, while the dense model runs out of memory from 12 images on.
Quality averages the five scores of Tab.~\ref{tab:minimax} and the three judge scores of Tab.~\ref{tab:qwen}, relative to the dense model.}
\label{fig:scaling}

\end{figure*}

\paragraph{Evaluation dataset.}
The test clips of our model experiments are randomly selected from a 2,000-video pool of VidGen-1M~\cite{tan2024vidgen}, where each comes with its ground truth and we prepare the condition modalities needed.
Inference mainly runs at 512\x768 with 81 frames.
Each experiment constructs its own cases and conditions from this set, namely 400 clips with 1 to 4 conditions in Tab.~\ref{tab:token_selection} and Tab.~\ref{tab:gap}, 200 clips with 1 to 5 conditions in Fig.~\ref{fig:stress}, 100 clips with 1 to 4 conditions in Tab.~\ref{tab:stress}, and 50 clips per condition in Fig.~\ref{fig:percond}.
Extension experiments are instead tested on benchmarks in their domain~\cite{li2026omnivbench, liu2025step1x, wu2026omnigen2, zhou2026pai, Motamed2025DoGV, dixit2026foleybench, khazatsky2024droid}.

\paragraph{Baselines.}
We compare popular token pruning-based methods.
The \textbf{Random} baseline drops tokens uniformly at random, the same share in every condition.
The \textbf{Reduced Resolution} baseline is the common way to reduce condition tokens widely used in controllable image and video generation~\cite{ben2026avcontrol, tan2026ominicontrol2, zhang2025easycontrol}, where every condition is encoded at one lower resolution.
The \textbf{DINO} attention baseline keeps the tokens that the CLS token of DINOv3~\cite{Simoni2025DINOv3} attends to most, which chooses the semantically most important tokens.
The \textbf{Saliency}~\cite{parihar2026keep} baseline applies DeepGaze~\cite{kummerer2025modeling} to keep the tokens with the highest predicted human-fixation density, which represents where a viewer would look.

\paragraph{Metrics.}
\begin{table}[t]
\centering

\caption{
\textbf{Perceptual Comparisons.}
Subscripts provide the change against the dense-conditioned teacher, and \method is consistent with the teacher on VBench~\cite{huang2024vbench}.
In the human study, about 90\% of the answers rate \method as equal or better in both condition following and quality.
}
\label{tab:stress}

\vspace{-2pt}
{\newcommand{\dlt}[1]{\textsubscript{\color{black!60}\fontsize{5.0}{6.0}\selectfont#1}}
\newcommand{\ovr}[2]{{\fontsize{5.6}{6.6}\selectfont #1}\,\raisebox{-0.35ex}{\fontsize{6.1}{7.1}\selectfont\boldmath\textbf{#2}}}
\tablestylesm{2.65pt}{1.515}\fontsize{6.5}{7.6}\selectfont
\begin{tabular}{c cccccc cc}
\toprule
& \multicolumn{6}{c}{VBench \up} & \multicolumn{2}{c}{Human study (\%)} \\
\cmidrule(lr){2-7}\cmidrule(l){8-9}
\makecell{\#\\cond.} & \makecell{Imaging\\quality} & \makecell{Aesth.\\quality} & \makecell{Motion\\smooth.} & \makecell{Temp.\\flicker.} & \makecell{Backgr.\\consist.} & \makecell{Overall\\consist.} &
\makecell{Equally\\follows?} & \makecell{Equally\\good?} \\
\midrule
1 & 73.4\dlt{\textminus0.2} & 54.1\dlt{+0.4} & 99.0\dlt{\textpm{}0.0} & 97.1\dlt{+0.1} & 94.4\dlt{+0.2} & 25.9\dlt{+0.1} & 84 & 95 \\
2 & 68.8\dlt{+0.1} & 54.9\dlt{+1.0} & 98.6\dlt{\textpm{}0.0} & 96.6\dlt{+0.1} & 93.9\dlt{+0.1} & 25.3\dlt{\textminus0.1} & 92 & 84 \\
3 & 65.8\dlt{+0.3} & 54.2\dlt{\textminus0.2} & 98.8\dlt{\textpm{}0.0} & 96.9\dlt{\textpm{}0.0} & 93.6\dlt{\textpm{}0.0} & 25.3\dlt{\textpm{}0.0} & 93 & 88 \\
4 & 66.7\dlt{\textminus0.3} & 53.9\dlt{+0.6} & 98.5\dlt{\textpm{}0.0} & 96.3\dlt{\textpm{}0.0} & 94.1\dlt{\textminus0.3} & 25.2\dlt{\textminus0.3} & 94 & 91 \\
\cmidrule(r){1-9}
{\boldmath\textbf{All}} & \ovr{67.7}{\textpm{}0.0} & \ovr{54.3}{+0.4} & \ovr{98.7}{\textpm{}0.0} & \ovr{96.7}{\textpm{}0.0} & \ovr{93.9}{\textpm{}0.0} & \ovr{25.4}{\textminus0.1} & \textbf{92} & \textbf{88} \\
\bottomrule
\end{tabular}
\par}
\end{table}

Perceptual quality and fidelity to the ground truth are measured with FID~\cite{heusel2017gans}, FVD~\cite{unterthiner2018towards}, and LPIPS~\cite{zhang2018unreasonable}.
Each condition also has its own metric, computed on the clips that carry it.
\textit{DINOv3}~\cite{Simoni2025DINOv3} cosine similarity is for identity references, \textit{mIoU} for segmentation, \textit{si-RMSE}~\cite{Eigen2014DepthMP} for depth, \textit{mean angular error} for surface normals, \textit{F1} for Canny edges, \textit{mPSNR} for inpainting and outpainting, \textit{PCK@0.1}~\cite{yang2012articulated} for human pose, \textit{TrajErr} for motion tracks, and the colour difference $\mathit{\Delta E}_{00}$ for grayscale.
We also select six dimensions of VBench~\cite{huang2024vbench} to measure videos without the ground truth.
More details are in the supplementary material.

\subsection{Token dropping and inference}
\label{sec:exp_token_selection}

\paragraph{Comparison of different token pruning methods.}
Tab.~\ref{tab:token_selection} compares the token pruning methods at the same overall budget, which drops 66\% of the condition tokens on average over all conditions.
Without training, codec pruning is the best training-free method in 8 of the 12 metrics, including \textit{FID}, \textit{FVD}, and \textit{LPIPS}.
After the same number of training steps, \method is the best fine-tuned model in 10 of the 12 metrics and stays close to the teacher.
Codec pruning selects the tokens in 1.13\,s per case on average on the CPU, in parallel with the VAE and text encoders, while DINO attention and saliency take 23.8\,s and 75.0\,s on the GPU.
The supplementary material provides a longer SFT training comparison, parallelism of modules, and per-condition token statistics of codec pruning.

\paragraph{Speedup and perceptual evaluation.}
\label{sec:exp_stress}
Fig.~\ref{fig:stress} measures the denoising time with 1 to 5 conditions per video.
Without the KV cache, the dense-conditioned teacher model slows down with every added condition, while ours grows much more slowly, where the speedup rises from 1.54\x\ to 2.30\x.
This gain is smaller than with full bidirectional attention, since our condition tokens never attend to the noise tokens, which also lets our base model support the KV cache setting, where their keys and values are computed once at the first denoising step and reused by every later step.
Ours is still 1.15\x\ to 1.37\x\ faster, and its KV cache is 62\% to 73\% smaller.

Tab.~\ref{tab:stress} compares ours with the teacher model on VBench~\cite{huang2024vbench} and a human study.
VBench shows that ours keeps the perceptual quality of the teacher model, with no score changing by more than 1.0 point.
In a blind A/B study, 20 raters took part and gave 384 answers.
They judged ours as good as or better than the dense model in 92\% of the answers for following the conditions and in 88\% for visual quality.

\begin{table}[t]

\caption{\textbf{Teacher--student gap ablation.}
We control the gap by the drop rate, dropping 90\% of student tokens for the larger gap and some teacher tokens for the smaller gap.
Both are worse than ours on at least 8 of the 9 metrics.
The best results are \textbf{highlighted}.}

\label{tab:gap}
\vspace{-3pt}

\tablestylesm{1.5pt}{1.05}
\resizebox{\columnwidth}{!}{%
\begin{tabular}{@{}l ccc cccccc@{}}
\toprule
& \multicolumn{3}{c}{Quality and Fidelity} & \multicolumn{6}{c}{Condition Following} \\
\cmidrule(lr){2-4}\cmidrule(l){5-10}
& FID & FVD & LPIPS & DINOv3 & si-RMSE & Edges F1 & mPSNR & PCK & TrajErr \\
& \dn & \dn & \dn & \up & \dn & \up & \up & \up & \dn \\
\midrule
Smaller gap & 23.7 & 123.9 & 0.290 & \textbf{0.290} & 0.793 & 0.406 & 20.26 & 0.853 & 12.04 \\
\cellcolor{ourgray}\textbf{Ours} & \cellcolor{ourgray}\textbf{23.0} & \cellcolor{ourgray}\textbf{113.8} & \cellcolor{ourgray}\textbf{0.278} & \cellcolor{ourgray}0.289 & \cellcolor{ourgray}\textbf{0.706} & \cellcolor{ourgray}\textbf{0.417} & \cellcolor{ourgray}\textbf{20.43} & \cellcolor{ourgray}\textbf{0.911} & \cellcolor{ourgray}\textbf{7.80} \\
Larger gap & 23.8 & 124.3 & 0.287 & 0.288 & 0.798 & 0.404 & 19.48 & 0.906 & 9.90 \\
\bottomrule
\end{tabular}}
\end{table}

\begin{table}[!tp]

\caption{\textbf{MiniMax-H3 Ref2VA on OmniVBench~\cite{li2026omnivbench}.}
Ours drops 71.6\% of condition tokens in the DiT and 36.2\% in the VLM.
It denoises 1.87\x\ faster, and no overall score drops more than 0.6 points.
Subscripts are changes from the dense model.}
\label{tab:minimax}
\vspace{-3pt}

\newcommand{\dlt}[1]{\textsubscript{\color{black!60}\fontsize{5.4}{6}\selectfont#1}}
\newcommand{\dltb}[1]{\textsubscript{\fontsize{6.6}{7}\selectfont\boldmath\textbf{#1}}}
\newcommand{\ovr}[2]{{\fontsize{5.6}{6}\selectfont #1}\,\raisebox{-0.35ex}{\fontsize{6.6}{7}\selectfont\boldmath\textbf{#2}}}
\tablestylesm{1.1pt}{1.0}\fontsize{6.1}{7.1}\selectfont
\begin{tabular}{l wc{21pt}wc{21pt}c@{\hskip 4pt}ccccc}
\toprule
& \multicolumn{2}{c}{Token Drop (\%)} & & \multicolumn{5}{c}{Quality \up} \\
\cmidrule(lr){2-3}\cmidrule(lr){5-9}
Task & DiT & VLM & \makecell{Denoise\\Speedup} & Technical & Aesthetic & Physical & \makecell{Reference\\Fidelity}
& \makecell{Instruction\\Realization} \\
\midrule
Content & 74.3 & 42.2 & 1.31\x & 20.1\dlt{\textminus0.1} & 51.1\dlt{\textminus0.3} & 64.7\dlt{+0.8} & 91.7\dlt{+0.4} & 95.6\dlt{+0.2} \\
Motion & 75.1 & 28.8 & 2.06\x & 15.3\dlt{\textminus0.2} & 50.4\dlt{\textminus0.3} & 63.1\dlt{\textminus1.2} & 53.9\dlt{+5.8} & 87.7\dlt{\textminus0.3} \\
Style & 44.3 & 19.1 & 1.13\x & 16.2\dlt{+0.2} & 54.1\dlt{\textminus0.4} & 65.4\dlt{+1.2} & 53.3\dlt{+5.0} & 84.8\dlt{\textminus0.7} \\
Structure & 79.7 & 42.3 & 2.21\x & 16.0\dlt{\textminus0.2} & 46.0\dlt{\textminus1.2} & 64.1\dlt{+0.7} & 86.3\dlt{+5.1} & 92.5\dlt{\textminus1.4} \\
Narrative & 71.8 & 25.0 & 1.78\x & 17.1\dlt{\textpm{}0.0} & 46.0\dlt{\textminus0.6} & 61.3\dlt{\textminus1.1} & 83.0\dlt{\textminus2.7} & 82.8\dlt{+3.6} \\
Multi-cont. & 77.9 & 47.2 & 2.02\x & 21.7\dlt{\textminus0.1} & 49.8\dlt{\textminus0.6} & 63.1\dlt{\textminus1.4} & 91.6\dlt{\textminus2.3} & 94.0\dlt{\textminus0.5} \\
Cross & 77.9 & 48.5 & 2.55\x & 19.2\dlt{+0.2} & 50.1\dlt{\textminus0.5} & 62.1\dlt{+0.8} & 77.2\dlt{\textminus0.4} & 93.1\dlt{+0.3} \\
\midrule
{\boldmath\textbf{Overall}} & {\boldmath\textbf{71.6}} & {\boldmath\textbf{36.2}} & {\boldmath\textbf{1.87\x}} & \ovr{17.9}{\textpm{}0.0} & \ovr{49.6}{\textminus0.6} & \ovr{63.4}{\textpm{}0.0} & \ovr{76.7}{+1.6} & \ovr{90.1}{+0.2} \\
\bottomrule
\end{tabular}
\vspace{-5pt}
\end{table}

\paragraph{Per-condition rate-distortion analysis.}
\label{sec:exp_drop_rate}
We also evaluate how each input modality alone affects quality and performance, with a single condition as the only input of the generation.
We then drop more and more tokens of this condition, starting with those of the lowest energy following Eq.~\ref{eq:gazer}.
Fig.~\ref{fig:percond} compares each output with the teacher.
The teacher's seed noise is $\pm2\sigma$ with $\sigma^2=(1+\frac{1}{S})\,\bar{v}/N$, where $\bar{v}$ is the mean variance of a case over $S=3$ teacher seeds and $N$ is the number of cases.
In Fig.~\ref{fig:percond}(c), dense conditions drop 26\% to 71\% of their tokens for free, while sparse conditions can drop 80\% to 94\%.
This shows that different condition modalities have different levels of robustness to token pruning.

\subsection{Teacher-student gap ablation}
\label{sec:exp_gap}

In \method, the teacher sees all condition tokens, while the student sees the tokens kept by codec pruning.
The dropped tokens form the gap between them.
For a smaller gap, the teacher is also pruned with half of the student's codec thresholds.
For a larger gap, the student drops 90\% of its condition tokens.
As shown in Tab.~\ref{tab:gap}, both variants are worse than ours on almost every metric, where FID rises from 23.0 to 23.7 and 23.8, and FVD from 113.8 to 123.9 and 124.3.
Therefore, the gap from codec pruning provides the best balance between what the teacher can teach and what the student can infer.

\subsection{Extend \method to other models}
\label{sec:exp_other_models}

We transfer codec pruning and \method together to three other models, which are MiniMax-H3 Ref2VA~\cite{minimax2026h3}, the image variant MiniMax-H3-Image~\cite{minimax2026h3image}, and Cosmos3~\cite{agarwal2026cosmos} for multi-task generation.

\subsubsection{Reference-to-video with MiniMax-H3 Ref2VA}
\label{sec:exp_minimax}
MiniMax-H3 Ref2VA~\cite{minimax2026h3} is a recent state-of-the-art video diffusion model with reference-to-video capabilities, i.e., the ability to be conditioned by images or videos besides text.
It can take up to 9 images and 3 videos as conditions.
Besides its diffusion transformer (DiT)~\cite{peebles2023scalable}, every condition image passes through the vision encoder (ViT) of its vision-language model (VLM)~\cite{bai2025qwen3}.
Condition videos are passed to this ViT as images at a 2 FPS rate.
The vision tokens of the VLM can amount to 33\% of the condition tokens in the DiT.
Therefore, we prune both kinds of tokens with the same codec signals in a single pass.

We evaluate on 237 selected OmniVBench~\cite{li2026omnivbench} cases, the reference-richest ones of each task, and present them by task family in Tab.~\ref{tab:minimax}.
Ours drops 71.6\% and 36.2\% of the DiT and VLM tokens, respectively, while denoising 1.87\x\ faster on average, with no overall score lower by more than 0.6 points.
In Fig.~\ref{fig:scaling} top, we show that our denoising speedup grows with the number of references, frames, and resolution. Adding \method can accelerate denoising by up to 3.72\x, while quality matches the teacher model and sometimes slightly exceeds it.
The data curation, the training details, and the results of \method on top of other acceleration techniques are in the supplementary material.

\subsubsection{Reference-to-image with MiniMax-H3-Image}
\label{sec:exp_h3image}
MiniMax-H3-Image~\cite{minimax2026h3image} supports text-to-image generation and editing with up to 16 reference images.
We evaluate on \mbox{GEdit-Bench-EN}~\cite{liu2025step1x} for image editing and OmniContext~\cite{wu2026omnigen2} for generation from multiple reference images, where codec pruning drops about half of the reference image tokens.
Since MiniMax-H3-Image requires a short side of at least 1024 pixels, we upscale smaller reference images to this size.
\method matches the teacher in Tab.~\ref{tab:qwen}, so the results stay consistent on existing benchmarks.

In Fig.~\ref{fig:scaling} bottom, the speedup grows with the number of reference images, from 1.36\x\ with 1 image to 2.99\x\ with 12, and with the resolution, from 2.25\x\ at $1024^2$ to 3.06\x\ at $2048^2$ pixels.
At $2048^2$ pixels, the dense model runs out of memory with 12 or more images even on a 141\,GB H200, while ours does not.
The data curation and training details are in the supplementary material.
\begin{table}[!tp]

\vspace{-2pt}
\caption{\textbf{MiniMax-H3-Image on GEdit-Bench-EN~\cite{liu2025step1x} and OmniContext~\cite{wu2026omnigen2}.}
Scores are VLM-judge ratings from 0 to 10, higher is better, and \textbf{Mean} averages them relative to the teacher.}
\label{tab:qwen}

\vspace{-1pt}
\tablestylesm{4.9pt}{1.1}
\newcommand{\mn}[1]{\cellcolor{black!7}#1}
\begin{tabular}{l c ccc c}
\toprule
Model & Ref. tokens & Instr.$\uparrow$ & Consist.$\uparrow$ & Quality$\uparrow$ & \mn{Mean (\%)} \\
\midrule
\grp{6}{GEdit-Bench-EN} \\
Teacher & 1{,}401 & 8.992 & 8.936 & 7.287 & \mn{100.0} \\
\ours +\method & 685 {\scriptsize(\textminus51\%)} & 9.061 & 8.868 & 7.295 & \mn{100.0} \\
\midrule
\grp{6}{OmniContext} \\
Teacher & 2{,}940 & 9.490 & 8.000 & 7.540 & \mn{100.0} \\
\ours +\method & 1{,}287 {\scriptsize(\textminus56\%)} & 9.527 & 8.015 & 7.560 & \mn{100.3} \\
\bottomrule
\end{tabular}
\end{table}

\subsubsection{Multi-task generation with Cosmos3}
\label{sec:exp_cosmos}
\begin{table}[!tp]

\vspace{-2pt}
\caption{\textbf{Comparison with the teacher on Cosmos3~\cite{agarwal2026cosmos}.}
Teacher values are the mean $\pm2\sigma$ over three seeds.
\method stays within this range or better on every task while denoising up to 2.30\x\ faster.
EE err.\ is the 3D end-effector position error.}
\label{tab:cosmos}
\vspace{-3pt}

\tablestylesm{0.8pt}{1.0}
\begin{tabular*}{\linewidth}{@{\extracolsep{\fill}} l l cc cc}
\toprule
& \multicolumn{3}{c}{\footnotesize Quality} & \multicolumn{2}{c}{\footnotesize Efficiency} \\[-2pt]
\cmidrule(lr){2-4}\cmidrule(lr){5-6}
\footnotesize Task & \footnotesize Metric & \makecell{\footnotesize Dense\\[-2pt]\footnotesize (teacher)} & \footnotesize +\method & \makecell{\footnotesize Token\\[-2pt]\footnotesize Drop {\fontsize{6}{7}\selectfont(\%)}} & \makecell{\footnotesize Denoise\\[-2pt]\footnotesize Speedup} \\
\midrule
Segmentation & mIoU\up & 0.715{\scriptsize$\pm$0.035} & 0.751 & 84.2 & 1.93\x \\
Blur & SSIM\up & 0.910{\scriptsize$\pm$0.004} & 0.911 & 45.0 & 1.36\x \\
Depth & si-RMSE\dn & 0.821{\scriptsize$\pm$0.096} & 0.846 & 84.7 & 1.92\x \\
Edge & F1\up & 0.462{\scriptsize$\pm$0.004} & 0.459 & 35.0 & 1.24\x \\
\midrule
Continuation & Physics-IQ\up & 50.6{\scriptsize$\pm$5.2} & 50.6 & 89.4 & 1.93\x \\
Video-to-audio & AV match\up & 0.864{\scriptsize$\pm$0.017} & 0.866 & 70.0 & 2.30\x \\
Video-to-action & EE err.\ (cm)\dn & 2.16{\scriptsize$\pm$0.06} & 2.21 & 44.0 & 1.21\x \\
\bottomrule
\end{tabular*}
\end{table}

Cosmos3~\cite{agarwal2026cosmos} is a Mixture-of-Transformers model for physical AI.
Its diffusion transformer can be conditioned on control videos such as edge, blur, depth, and segmentation, on history frames for video continuation, and on an input video for audio and action generation.
All condition tokens join the generated tokens in one sequence, so codec pruning applies without modification.
We use Cosmos3-Nano, which has 16B parameters.
Tab.~\ref{tab:cosmos} tests four of its conditioned tasks.
On control-to-video with 20 selected PAI-Bench-C~\cite{zhou2026pai} cases per control, \method stays within the teacher's $\pm2\sigma$ seed noise or better, with 84\% of the segmentation and depth tokens dropped for 1.9\x\ faster denoising and lower drops of 45\% and 35\% for blur and edge.
On continuation of the 66 center-view Physics-IQ videos~\cite{Motamed2025DoGV}, it drops 89\% of the given-frame tokens, denoises 1.93\x\ faster, and keeps the dense Physics-IQ score.
On video-to-audio with 50 selected FoleyBench~\cite{dixit2026foleybench} clips, it drops 70\% of the video tokens and denoises 2.30\x\ faster at the same AV match, the probability that a multimodal judge~\cite{Xu2025Qwen3OmniTR} accepts the soundtrack.
On video-to-action with 50 randomly selected DROID~\cite{khazatsky2024droid} episodes, the official setting reads only 17 input frames, but \method still drops 44\% of their tokens, denoises 1.21\x\ faster, and its 3D end-effector position error against the robot log stays close to the teacher's.
Visualizations and details are in the supplementary material.

\FloatBarrier

\section{Conclusion}
\label{sec:conclusion}

In \method, codec pruning drops redundant condition tokens with a standard video codec, and on-policy self-distillation adapts the model to the kept tokens.
\method matches the dense teacher while dropping 66\% of the condition tokens on average.
It works across \numcond condition types of different modalities and extends to MiniMax-H3 Ref2VA, MiniMax-H3-Image, and Cosmos3.
A promising direction for future work is to bring codec pruning to a finer granularity, for example by reducing the channels of each kept token according to its codec residual.

\section{Acknowledgments}
\label{sec:ack}
The authors acknowledge the University of Virginia Research Computing and Data Analytics Center (Rivanna).

{
    \small
    \bibliographystyle{ieeenat_fullname}
    \bibliography{main}
}

\clearpage
\setcounter{page}{1}
\maketitlesupplementary
\raggedbottom
\renewcommand{\thesection}{\Alph{section}}
\renewcommand{\theHsection}{S\Alph{section}}
\renewcommand{\thepage}{S\arabic{page}}
\setcounter{section}{0}
\renewcommand{\thetable}{S\arabic{table}}
\setcounter{table}{0}
\renewcommand{\thefigure}{S\arabic{figure}}
\renewcommand{\theHfigure}{S\arabic{figure}}
\setcounter{figure}{0}

In this supplementary material, we provide additional details and results for the main paper, which are organized as follows.
\begin{itemize}
\setlength{\itemsep}{1pt}\setlength{\parskip}{0pt}\setlength{\parsep}{0pt}
\item Supplement~\ref{sec:supp_impl} provides the implementation details of our multi-condition video model, built on a pretrained 22B DiT.
\item Supplement~\ref{sec:supp_codec} provides the implementation details of our video codec.
\item Supplement~\ref{sec:supp_data} provides the test data, metrics, and settings of all experiments.
\item Supplement~\ref{sec:supp_more_quant} provides additional quantitative results of our multi-condition video model.
\item Supplement~\ref{sec:supp_qual} provides qualitative results of our multi-condition video model.
\item Supplement~\ref{sec:supp_ablation_more} provides additional ablation studies.
\item Supplement~\ref{sec:supp_minimax} provides additional details and results on MiniMax-H3 Ref2VA~\cite{minimax2026h3}.
\item Supplement~\ref{sec:supp_h3img} provides additional details and results on MiniMax-H3-Image~\cite{minimax2026h3image}.
\item Supplement~\ref{sec:supp_cosmos} provides additional details and results on Cosmos3-Nano~\cite{agarwal2026cosmos}.
\item Supplement~\ref{sec:supp_limitations} discusses the limitations of our method.
\end{itemize}

\section{Implementation details}
\label{sec:supp_impl}

This section extends Sec.~\ref{sec:exp_setup} of the main paper with further implementation details.

\begin{figure*}[t]
\centering
\includegraphics[width=\textwidth]{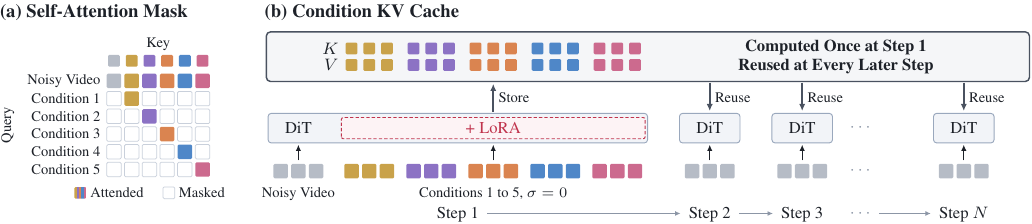}

\caption{\textbf{Architecture of our multi-condition model.}
(a) Noise tokens attend to all tokens, while each clean condition attends to its own tokens.
(b) Therefore, the keys and values of the conditions are computed once with our LoRA at step 1 and reused in all later steps, where the noise tokens alone enter the DiT.}

\label{fig:supp_arch}
\end{figure*}

\paragraph{Our internal model.}
\label{sec:supp_base}
Our internal model is a pretrained DiT with 22B parameters (Tab.~\ref{tab:supp_base}), and Fig.~\ref{fig:supp_arch} shows how our multi-condition model uses it.
Its 6B audio branch is not loaded, which leaves 16B parameters for video-only generation in all of our results.
Each condition token enters the DiT with $\sigma=0$ and attends to the tokens of its own condition.
Therefore, its keys and values are computed once and cached~\cite{tan2026ominicontrol2}.
Our LoRA~\cite{hu2021lora} is applied to the condition tokens.
\label{sec:supp_distilled}
Our internal model also has an 8-step distilled version~\cite{yin2024one} with the same architecture, which supports the KV cache in the same way.
The dense LoRA trained on our internal model works directly on this distilled version, so we use it as the teacher without further training.
The student starts from the same LoRA and is trained with \method for 2K steps, where every denoising step of the distilled version is supervised.
We adopt this setting for the main example in Fig.~\ref{fig:teaser}, the speedup test in Fig.~\ref{fig:stress}, and the privileged teacher study in Supp.~\ref{sec:supp_privileged}.

\begin{table}[t]

\caption{\textbf{Our internal model.}
We list the components of our internal model, whose audio branch is not loaded.}
\label{tab:supp_base}

\tablestyle{5pt}{1.1}
\begin{tabular}{l l}
\toprule
Component & Setting \\
\midrule
Transformer & 48 blocks, width 4{,}096 (32 heads \x 128) \\
Text & 12B LLM encoder, 8-layer connector, cross-attention \\
VAE & 32\x\ spatial, 8\x\ temporal compression \\
Tokens & one per latent cell (1\x1\x1 patches), 128 channels \\
Positions & 3D RoPE~\cite{su2024roformer} (time, height, width); after pruning, \\
& a token keeps its position in the full grid \\
Parameters & 22B in the checkpoint, 21B in the \\
& transformer and 1B in the text projection \\
Audio branch & audio stream and audio--video cross-attention, \\
& 6B of the transformer, not loaded \\
Loaded & 15B in the transformer and 1B in the \\
& text projection, 16B in total \\
\bottomrule
\end{tabular}
\end{table}

\paragraph{Training.}
In Stage 1, we train the dense model as a LoRA of rank 320 with about 2B parameters on top of our internal model.
This stage uses all condition tokens and runs for 50,000 steps on 48 GPUs with a global batch size of 48.
In Stage 2, each fine-tuned row of Tab.~\ref{tab:token_selection} starts from the dense model and is trained for 2K steps on the same data, with 49 frames at 480\x640 and a global batch size of 16.
The SFT baselines continue to train the dense LoRA with the flow matching loss~\cite{lipman2022flow} on the ground-truth clips, each with its own token selection method.
For \method, we copy the dense LoRA into a trainable student and freeze the dense model as the teacher.
Both the student rollout and the teacher target use a single conditional prediction without classifier-free guidance (CFG)~\cite{ho2022classifier}, while CFG is applied at inference.
\method is optimized with AdamW at a learning rate of $10^{-4}$.
During training, the student receives the tokens that remain after threshold-based pruning with the fixed thresholds of Tab.~\ref{tab:supp_codec_settings}.
Across the four base models, our training data is drawn from parts of OpenVid-1M~\cite{nan2025openvid}, VidGen-1M~\cite{tan2024vidgen}, SpatialVID~\cite{wang2026spatialvid}, DL3DV-10K~\cite{ling2024dl3dv}, and DROID~\cite{khazatsky2024droid}, together with images and additional clips from our internal data pool.

\paragraph{Conditions.}
Besides the text prompt, each training sample draws 1 to 5 of the \numcond condition modalities described in Sec.~\ref{sec:exp_setup}.
Their labels are extracted offline from each clip, except for edge, blurring, and grayscale, which are computed on the fly during data loading.

Object-level conditions include segmentation, bounding boxes, and identity.
For segmentation, Qwen3-VL~\cite{bai2025qwen3} locates the objects in the first frame with bounding boxes, from which SAM3~\cite{carion2026sam} tracks each object through the clip.
The resulting masks are rendered in one colour per object, either filled or as outlines.
Bounding boxes form a separate modality and are rendered from the same masks.
For identity, we create a reference image of each object from its crop, either as a single-view identity, which shows one view of the object, or as a multi-view identity, which shows several views of the object in one image.

Structural conditions include sketch, edge, depth, surface normals, and human pose.
For sketch, we use either the line drawings of Informative Drawings~\cite{chan2022learning} in the anime, contour, and open sketch styles, or the straight lines of M-LSD~\cite{gu2022towards}.
Edge is a separate modality that uses Canny edges with random thresholds.
Depth maps are predicted by the small model of Video Depth Anything~\cite{chen2025video} and normalized per clip without further smoothing, while surface normals are predicted by NormalCrafter~\cite{bin2025normalcrafter}.
For human pose, we render the whole-body keypoints of Sapiens-1B~\cite{khirodkar2024sapiens} as a skeleton for each person tracked by SAM3.

Motion conditions include sparse and dense tracking.
For tracking, CoTracker3~\cite{karaev2025cotracker3} tracks a grid of points through each clip.
Sparse tracking, denoted as motion track in Tab.~\ref{tab:token_selection} and Fig.~\ref{fig:percond}, renders 1 to 20 of the fastest points as coloured squares.
Dense tracking instead renders all points, coloured by their first-frame position and depth as in Diffusion as Shader~\cite{Gu2025DiffusionAS}.

Pixel-level conditions include blurring, grayscale, frame interpolation, continuation, inpainting, and outpainting.
For blurring, 70\% of the clips are degraded by a Gaussian blur with the kernel settings of Cosmos-Transfer1~\cite{alhaija2025cosmos} followed by down-up sampling, and the rest by a disk, Gaussian, or box kernel, from which the model restores the sharp clip.
For grayscale, the condition retains the luma of the clip, which the model then colourizes.
For frame interpolation and continuation, we retain the first frame, the first and last frames, or other frames of the clip and fill the remaining frames with grey.
Inpainting masks follow DeepFillv2~\cite{yu2019free} and LaMa~\cite{suvorov2022resolution} or are obtained from SAM3, while outpainting masks follow M3DDM~\cite{fan2023hierarchical} and cover either the whole clip or the first frame alone.

\section{Video codec implementation details}
\label{sec:supp_codec}

This section extends Secs.~\ref{sec:prelim} and~\ref{sec:codec_token_mask} of the main paper with the details of token pruning, the codec reader and its fixes, and the H.264~\cite{wiegand2003overview} encoding settings.

\begin{figure*}[t]
  \centering
  \includegraphics[width=0.88\textwidth]{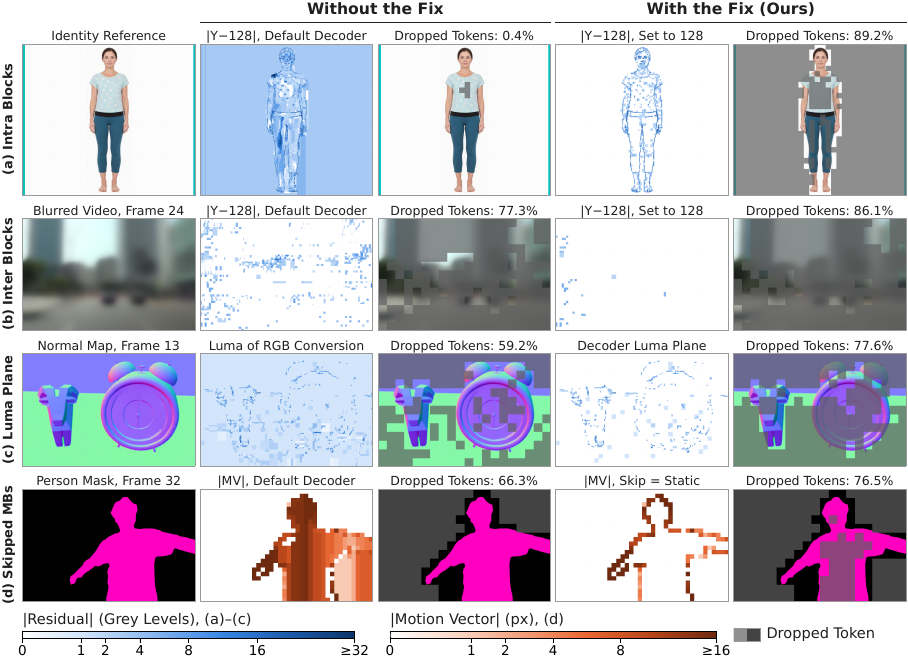}

  \caption{\textbf{Codec fixes.}
  Each row shows an input, followed by its codec signal and the tokens after codec pruning, without the corresponding fix on the left and with it on the right.
  The four fixes (a) set uncoded intra blocks to $128$, (b) set uncoded inter blocks to $128$, (c) read the residual from the luma plane of the decoder instead of an RGB conversion, and (d) treat skipped macroblocks as static.
  Without these fixes, flat regions and backgrounds are mistaken for new content, and their tokens are not dropped.
  }
  \label{fig:supp_codec}

\end{figure*}

\subsection{Token pruning}
\label{sec:supp_pruning}

This subsection details the two pruning mechanisms introduced in Sec.~\ref{sec:codec_token_mask}.

\paragraph{Rescaling with the condition size.}
Since motion vectors are measured in pixels, the same movement yields a larger motion score at a higher resolution.
In contrast, the residual score is averaged over the pixels of each token and decreases when an image is enlarged, since the same detail then spreads over more pixels.
Therefore, we define the thresholds of Eq.~\ref{eq:admissible} at a standard size and rescale them by the condition size $H\times W$, which can be formulated as:
\begin{equation}
  \tau_m=\tau_m^0\sqrt{\tfrac{HW}{H_0W_0}},\quad\tau_r=\tau_r^0\sqrt{\tfrac{s_0}{\max(s_0,\min(H,W))}},
  \label{eq:thresholds}
\end{equation}
where $\tau_m^0$ and $\tau_r^0$ are the base thresholds listed in Tab.~\ref{tab:supp_codec_settings}, defined at the standard size $H_0\times W_0=480\times640$ and the standard short side $s_0=1{,}024$, respectively.
The base thresholds are fixed for each condition type, and the rescaling adapts them to the resolution of the condition.
The residual rescaling applies to images with a short side above $s_0$.

\paragraph{Priority-based sorting.}
Priority-based sorting ranks the tokens by the energy $E_i=\bar R_i\,(1+\lambda\bar M_i)$ of Eq.~\ref{eq:gazer}, where $\bar R_i$ and $\bar M_i$ are the percentile ranks of $R_i$ and $M_i$ among the tokens of the same condition, both in $[0,1]$, and $\lambda$ weights the motion.
In practice, we set $\lambda=1$.
The tokens with the lowest energy are dropped first until the target drop rate is reached.
Since motion scales the energy of a token by a factor between $1$ and $1+\lambda$, tokens with a low residual tend to be dropped first, as the codec already predicts their content.
When priority-based sorting follows threshold-based pruning, the ranks are computed among the tokens with $b_i=1$.

\subsection{Codec reader and fixes}
\label{sec:supp_reader}

\paragraph{Choice of H.264.}
H.264~\cite{wiegand2003overview} partitions each frame into macroblocks of $16\times16$ pixels, so each token covers whole macroblocks, and its scores follow directly from them.
In contrast, H.265~\cite{Sullivan2012OverviewOT} and AV1~\cite{Han2020ATO} select blocks of up to $64\times64$ and $128\times128$ pixels, respectively, through rate-distortion optimization.
A single block can thus span several tokens, which makes the token scores depend on the partition decisions of the encoder.
AV1 also predicts with warped and overlapped block motion, so the motion of a block can no longer be described by a single displacement.
H.264 is also the fastest to encode.
At the fastest setting of each encoder, x264 encodes a 49-frame $480\times640$ clip in 0.06\,s, while x265 and the libaom AV1 encoder take 0.17\,s and 0.65\,s, respectively.

\paragraph{Codec reader.}
Our codec reader builds upon the H.264 decoder of OneVision-Encoder~\cite{tang2026onevision}, which is patched to output the residual instead of the reconstructed frame.
While its default behaviour is suitable for reconstruction, it mistakes flat backgrounds and the interior of moving shapes for new content, and these false positives lower the drop rate.
As shown in Fig.~\ref{fig:supp_codec}, we remove these false positives with four fixes that apply to all conditions.

\paragraph{(a) Uncoded intra blocks.}
An intra block without coded coefficients holds its prediction, which extrapolates the residual of neighbouring blocks across flat regions.
We set these blocks to $128$, which denotes a zero residual, and this increases the drop rate of identity references from 14.4\% to 74.8\%.

\paragraph{(b) Uncoded inter blocks.}
An inter block without coded coefficients, including a skipped block, holds a motion-compensated prediction that copies the residual of its reference frame, so the residual of earlier frames persists.
We also set these blocks to $128$, which increases the drop rate of condition videos from 70.4\% to 77.0\%.

\paragraph{(c) Luma plane.}
We read the residual directly from the luma plane of the decoder.
Converting the decoded frame to RGB would instead interpret its 4:4:4 chroma as 4:2:0, which shifts the zero level away from $128$ and introduces false residual.

\paragraph{(d) Skipped macroblocks.}
A skipped macroblock carries the motion vector predicted from its neighbours, so the flat interior of a moving mask appears to be moving.
We export a skip flag from the decoder and treat skipped macroblocks as static, which increases the drop rate of condition videos from 74.6\% to 77.0\%.

\subsection{H.264 encoding settings}
\label{sec:supp_h264}

In our simplified setting, an H.264 encoder codes the first frame of a condition as an I-frame from its own pixels and each subsequent frame as a P-frame.
Each P-frame is predicted from a reference frame with one motion vector per block, and the remaining prediction error is coded as the residual.
The constant rate factor (CRF) controls how coarsely this residual is quantized, where a higher CRF removes more small residuals.
The preset controls how much effort the encoder spends on its decisions, such as the motion search and the number of reference frames.

\paragraph{Presets.}
We adopt the fast presets in Tab.~\ref{tab:supp_codec_settings} for two reasons.
First, they encode quickly, so codec pruning finishes before the GPU work on average for all four base models, as shown in Tab.~\ref{tab:supp_prep}.
Second, they search a small motion range over a single reference, which is the previous frame.
In contrast, slower presets search up to 16 earlier frames, so their motion vectors can point many frames back and no longer measure the motion since the previous frame.
On a test set of 27 condition videos, our setting encodes a clip in 0.058\,s, while the medium and veryslow presets take 5.24\x\ and 21.5\x\ as long, respectively.
When x264 is allowed to search 16 frames in the history, 34.5\% of the coded P-frame blocks refer to a frame that is 2 or more frames back.

\paragraph{B-frames.}
We disable B-frames, which predict blocks from both earlier and later frames.
Since a B-frame block can take its motion from a later frame or average two predictions, its motion vector and residual no longer describe the change since the previous frame.
In contrast, with P-frames alone, every motion vector points to the previous frame.
With the medium preset, 48.3 of the 70.3 frames in a condition video become B-frames on average, and 54.5\% of their coded blocks are predicted from a later frame.

\paragraph{Encoder settings.}
Each condition is converted from 8-bit RGB to YUV 4:4:4 and encoded once.
We use CRF 33 for videos and CRF 35 for images, such that small prediction errors leave no residual.
A condition of up to 250 frames consists of one I-frame followed by P-frames.
We disable adaptive quantization and the deblocking filter of the decoder, which would otherwise introduce residual in flat regions and spread it across block edges.
Since the decisions of the encoder depend on the number of threads, we fix it to one thread per two macroblock rows.

\begin{table}[t]

\caption{\textbf{Codec settings.}
The upper part lists the settings shared by all conditions, and the lower part lists those that vary across condition types.
The motion threshold $\tau_m^0$ and the residual threshold $\tau_r^0$ of Eq.~\ref{eq:thresholds} are measured in pixels at 480\x640 and in grey levels, respectively.
Lower thresholds drop fewer tokens but retain small changes that carry content, so we adopt them for motion tracks and boxes.
However, thresholds that are too low lead to a small drop rate.}
\label{tab:supp_codec_settings}

\tablestylesm{3pt}{1.1}
\begin{tabular}{@{}>{\raggedright\arraybackslash}p{0.47\columnwidth} c c c c@{}}
\toprule
\multicolumn{5}{@{}l}{\textit{Shared by all conditions}} \\
Encoder & \multicolumn{4}{l@{}}{x264, 8-bit YUV 4:4:4 input} \\
Frame types & \multicolumn{4}{l@{}}{one I-frame, then P-frames} \\
Key-frame interval & \multicolumn{4}{l@{}}{250 frames, no scene cut} \\
Reference frames & \multicolumn{4}{l@{}}{1, the previous frame} \\
Adaptive quantization & \multicolumn{4}{l@{}}{off} \\
Deblocking & \multicolumn{4}{l@{}}{off when decoding} \\
Threads & \multicolumn{4}{l@{}}{one per two macroblock rows} \\
\midrule
Condition & Preset & CRF & $\tau_m^0$ & $\tau_r^0$ \\
\midrule
Segmentation, human pose, depth, normal, sketch, edges, dense track, grayscale, interpolation and continuation, inpainting, outpainting, first-frame outpainting & ultrafast & 33 & 5 & 1.0 \\
Motion track & ultrafast & 33 & 2 & 0.5 \\
Boxes & ultrafast & 33 & 1 & 0.1 \\
Blur & veryfast & 33 & 5 & 1.0 \\
Identity & superfast & 35 & -- & 1.5 \\
\bottomrule
\end{tabular}
\end{table}

\paragraph{Per-condition settings.}
As listed in Tab.~\ref{tab:supp_codec_settings}, some settings are shared by all conditions, while others vary across condition types.
Tab.~\ref{tab:supp_alloc} reports the share of tokens dropped for each condition.
The thresholds control the trade-off between the drop rate and the content that the student has to recover.
Higher thresholds drop more tokens, including those with small new content that is harder to learn, while lower thresholds retain this content at the cost of a lower drop rate.
Most conditions share the ultrafast preset, CRF 33, $\tau_m^0=5$ pixels at $H_0\times W_0=480\times640$, and $\tau_r^0=1.0$ grey level in Eq.~\ref{eq:thresholds}.
Motion tracks and bounding boxes consist of a few small flat shapes, where even a small motion or residual indicates real content.
Therefore, we adopt lower thresholds for them, with 2 pixels and 0.5 for motion tracks and 1 pixel and 0.1 for bounding boxes.
Blurred conditions are filtered real videos, on which the coarse motion search of ultrafast produces noisy motion vectors that lower the drop rate.
Therefore, we use veryfast for blurring.
Identity references are encoded as a single I-frame without motion, so their pruning is determined by the residual threshold alone, with $\tau_r^0=1.5$ at a short side of $s_0=1{,}024$ pixels.

\section{Evaluation setup and metrics}
\label{sec:supp_data}
\label{sec:supp_metrics}

This section provides the test data, metrics, and settings of all experiments, organized by base model.

\subsection{Our multi-condition video model}
\label{sec:supp_eval_ours}

\subsubsection{Sampling and pruning}
We sample our multi-condition model with 30 steps and a CFG scale of 3 in Tabs.~\ref{tab:token_selection} to~\ref{tab:gap} and Fig.~\ref{fig:percond}, while Fig.~\ref{fig:stress} uses its distilled version.
We apply threshold-based pruning with the thresholds in Tab.~\ref{tab:supp_codec_settings} to the Codec (ours) rows of Tab.~\ref{tab:token_selection}, as well as in Tabs.~\ref{tab:stress} and~\ref{tab:gap} and Figs.~\ref{fig:stress} and~\ref{fig:supp_qual_ours}.
In contrast, Fig.~\ref{fig:percond} drops a fixed share of the tokens of each condition through priority-based sorting over all of its tokens.

\subsubsection{Metrics}
\paragraph{Video quality.}
FID~\cite{heusel2017gans} is computed on up to 10,000 output frames sampled uniformly from each clip, and FVD~\cite{unterthiner2018towards} on 49 uniformly sampled frames per clip, both against the ground truth.
LPIPS~\cite{zhang2018unreasonable} compares each output frame with the corresponding ground-truth frame.
Before scoring, the frames are resized to 256\x384 for FID and FVD and to 480\x640 for LPIPS.
In Tab.~\ref{tab:stress}, imaging quality is measured by the MUSIQ~\cite{ke2021musiq} score in VBench~\cite{huang2024vbench}.
\paragraph{Condition following.}
Each condition metric is computed on the clips that carry this condition.
For most conditions, we re-extract the condition from the output with the same extractor that produced the input.
For identity, we compute the cosine similarity between the DINOv3~\cite{Simoni2025DINOv3} ViT-B/16 class tokens of each identity reference and those of sampled output frames.
For segmentation, SAM3~\cite{carion2026sam} tracks each condition object in the output from its first-frame box, and mIoU averages the IoU over the condition objects.
For depth, si-RMSE~\cite{Eigen2014DepthMP} compares the Video Depth Anything~\cite{chen2025video} depth maps of the output and the ground truth, while AbsRel in Fig.~\ref{fig:percond} is computed after aligning the scale and shift per video.
For surface normals, we measure the mean angle in degrees between the NormalCrafter~\cite{bin2025normalcrafter} normals of the output and those of the ground truth.
For edge, we compute the pixel-wise F1 score between the Canny edges of the output and those of the ground truth.
For sketch, we compute the SSIM between the Informative Drawings~\cite{chan2022learning} contour drawings of the output and those of the ground truth.
For bounding boxes, we compute the AP50 of class-agnostic YOLOv8m detections on the output, which are compared with the condition boxes frame by frame.
For blurring, denoted as Deblur in Fig.~\ref{fig:percond}, we compute the PSNR of the output against the ground truth.
For inpainting and outpainting, mPSNR is the PSNR over the generated region, i.e., the pixels that the condition leaves empty, and thus excludes the given pixels.
For human pose, we compute PCK@0.1~\cite{yang2012articulated} of the Sapiens-1B~\cite{khirodkar2024sapiens} body keypoints of the people in the output against those of the ground truth.
For motion tracks, TrajErr is the mean distance in pixels at 480\x640 between the requested point tracks and the same points re-tracked on the output with CoTracker3~\cite{karaev2025cotracker3}.
For grayscale, $\Delta E_{00}$ is the CIEDE2000 colour difference between the output and the ground truth.
Dense tracking, frame interpolation, continuation, and first-frame outpainting have no dedicated metric and are evaluated through FID, FVD, and LPIPS.
The metrics for sketch, bounding boxes, and blurring are reported in Fig.~\ref{fig:percond}, and the other condition metrics are reported in Tab.~\ref{tab:token_selection}.

\subsubsection{Test cases with more conditions}
Each test case in Fig.~\ref{fig:stress} uses a different test video, and its conditions are drawn at random from the valid combinations of the training pool, without filtering by quality or results.
In these cases, the drop rate depends more on the condition types of a case than on their number, since dense appearance conditions drop 50--63\% of their tokens, while sparse conditions such as motion tracks and bounding boxes drop about 96\%.
With the KV cache, the keys and values of the condition tokens are reused at every step, so pruning shortens the attention from the noise tokens to these tokens.
Without the cache, every step additionally processes all of these condition tokens.

\subsection{MiniMax-H3 Ref2VA}
\label{sec:supp_eval_minimax}

\subsubsection{Evaluation dataset details}
Tab.~\ref{tab:minimax} is evaluated on 237 cases selected from the 813 cases of OmniVBench~\cite{li2026omnivbench}, which cover all seven task families of the benchmark.
Since each case requires a 50-step generation by both the dense model and \method, we focus on cases with long input sequences, which are the main target of codec pruning.
Specifically, we rank the cases within each task ID of the benchmark by their number of reference tokens in the DiT and the VLM, as estimated from the inputs.
For task families with one, two, and three or more task IDs, we select the top 30, 15, and 10 cases per task ID, respectively, where ties are resolved in favour of less-covered categories.
From this selection, we evaluate the cases that both models complete on a single H200 GPU.

For prompt preparation, MiniMax recommends converting the instruction and references into a detailed prompt, with either its online service H3-Context-IR, as used by OmniVBench, or a system built on its official prompting guide.
We adopt the latter approach and rewrite the prompts with Qwen3-VL-32B~\cite{bai2025qwen3}, whose system prompt is this guide adapted to silent videos.
Both models generate videos with the same rewritten prompts and the same seed, while the judge receives the original instructions of the benchmark.

\subsubsection{Metrics}
For video quality, we follow the official code of OmniVBench, where technical quality is measured by the DOVER++~\cite{Wu2022ExploringVQ} technical score, aesthetic quality by the Aesthetic Predictor V2.5 score, and physical plausibility by the physics score of UnifiedReward 2.0~\cite{Wang2025UnifiedRM}.
Reference fidelity and instruction realization follow the official prompts and rubrics and are rescaled to 0--100.
The benchmark judges these two metrics with Gemini 3.1 Pro, to which we have no API access.
Therefore, we adopt the open Qwen3.8-Flash-Next~\cite{Qiu2026OnTD} as the judge with the same prompts, a large vision-language model with 125B parameters, 6B of which are activated, that natively supports video input.
We observe that it provides a reasonable understanding of the generated videos for these rubrics.
Token drop is defined as one minus the ratio of remaining to dense condition tokens in each case, which is measured before the DiT and, for the reference tokens in the VLM, before its language model.
Speedup is defined as the summed denoising time of the dense model divided by that of \method.
Both are computed per task family, and Overall is the mean over the seven families, following the aggregation of OmniVBench.
In Fig.~\ref{fig:scaling} top, quality is the ratio of \method to the dense model, averaged over the five metrics.
Each video is scored at a 480-pixel short side, so that all resolutions are evaluated consistently.

\subsubsection{Inference settings}
We use the released 50-step sampler for Tab.~\ref{tab:minimax} and the 8-step FastH3-Omni of FastVideo~\cite{pmlr-v267-zhang25m} for both the speed and quality measurements in Fig.~\ref{fig:scaling} top.
In the DiT, we apply threshold-based pruning to the references, and for reference videos, we further drop part of the remaining tokens through priority-based sorting.
In the VLM, we drop the vision tokens with $b_i=0$ for reference videos.
For reference images, we drop the vision tokens that are dropped in the DiT and also fall below a quarter of the thresholds.
We do not use or compare with a KV cache on MiniMax-H3 Ref2VA, since its reference tokens attend to the noise tokens, and we observe that caching their keys and values across denoising steps degrades condition following for both reference images and videos.

\subsection{MiniMax-H3-Image}
\label{sec:supp_eval_h3img}

\subsubsection{Evaluation dataset details}
Tab.~\ref{tab:qwen} is evaluated on all 606 cases of GEdit-Bench-EN~\cite{liu2025step1x} and all 400 cases of OmniContext~\cite{wu2026omnigen2}.

\subsubsection{Metrics}
The judges of these benchmarks are accessed through APIs.
For reproducibility, we instead use the open Qwen3.8-27B~\cite{Qiu2026OnTD} with the official prompts to rate Instr., Consist., and Quality from 0 to 10, so our scores are not comparable to those of the original judges.
Instr.\ measures editing success on GEdit-Bench and prompt following on OmniContext.
Consist.\ measures the absence of over-editing on GEdit-Bench and the consistency of subjects and scenes with the references on OmniContext.
Quality is the lower of the naturalness and artifact scores, both of which are rated on the output alone.

\subsubsection{Inference settings}
All cases use the official prompt enhancer and the released 6-step sampler.
In Tab.~\ref{tab:qwen} and Fig.~\ref{fig:scaling} bottom, we apply threshold-based pruning in the DiT and the image pruning rule of MiniMax-H3 Ref2VA in the VLM.
In Figs.~\ref{fig:supp_trainfree_h3img} and~\ref{fig:supp_addon_h3img}, the VLM instead drops the vision tokens that are dropped in the DiT.

\subsection{Cosmos3-Nano}
\label{sec:supp_eval_cosmos}

In Tab.~\ref{tab:cosmos}, the values of the dense model are the mean $\pm2\sigma$ over three seeds, and the drop rates of priority-based sorting are selected on the test cases of the table.
The denoising speedups in Tab.~\ref{tab:cosmos} are measured on one A100 GPU with the released inference pipeline.
Note that we observe that the measured speedups also vary with the GPU and inference optimizations.
For example, on faster GPUs, the short video-to-action workload is limited by kernel launch overhead rather than computation.

\subsubsection{Transfer}
Control-to-video generates 121 frames from a single control at 720\x1280, or at 832\x1104 for 4:3 clips, and we evaluate 20 selected PAI-Bench-C~\cite{zhou2026pai} cases per control.
We use the official scorer of PAI-Bench-C, which compares the output with the source video through SSIM for blur, Canny F1 for edge, si-RMSE~\cite{Eigen2014DepthMP} of estimated depth for depth, and mIoU of SAM 2~\cite{Ravi2024SAM2S} masks for segmentation.
These scores are not comparable to the metrics of our multi-condition model, which re-extract each condition with our own extractors and evaluate blurring with PSNR.
Segmentation and depth use threshold-based pruning.
For blur and edge, threshold-based pruning moves the quality outside the range of the teacher, so we instead adopt priority-based sorting at the highest tested drop rate that remains within this range, which is 45\% for blur and 35\% for edge.

\subsubsection{Continuation}
For continuation, the model observes the first 73 frames of each of the 66 center-view Physics-IQ~\cite{Motamed2025DoGV} videos and generates the next 120 frames, with threshold-based pruning on the given frames.
The results are evaluated with the Physics-IQ score.

\subsubsection{Video-to-audio}
For video-to-audio, we generate soundtracks for 50 selected FoleyBench~\cite{dixit2026foleybench} clips.
AV match is the probability that a Qwen3-Omni~\cite{Xu2025Qwen3OmniTR} judge considers the soundtrack to match the video, which is 0.842 for the ground truth.
Threshold-based pruning drops 50\% of the video tokens, while priority-based sorting at a fixed drop rate of 70\% drops more and remains within the range of the teacher, as does the tested rate of 95\%.

\subsubsection{Video-to-action}
For video-to-action, the model reads 17 frames from three DROID~\cite{khazatsky2024droid} camera views and predicts the 16 end-effector actions between them, with threshold-based pruning.
We evaluate this task on 50 randomly selected episodes.
EE err.\ is the distance between the end-effector position after the 16 predicted actions and the position recorded in the robot log.

\begin{table*}[t]

\caption{\textbf{Codec pruning finishes within the GPU encoding time, even on a modest CPU.}
We report the mean seconds per case on one A100 GPU with 6 CPU cores, over the 400 test clips of Tab.~\ref{tab:token_selection} for our internal model and 100 cases for each of the other models.
GPU work refers to VAE encoding, together with text encoding for our internal model, and extra time left is the GPU work time minus the codec pruning time.}
\label{tab:supp_prep}

\tablestylesm{4.4pt}{1.1}
\begin{tabular}{l l ccc}
\toprule
Model & Input per case & \makecell{GPU work\\(s)} & \makecell{Codec\\pruning (s)} & \makecell{Extra time\\left (s)} \\
\midrule
Our internal model & 1--4 conditions, 81 frames, 512\x768 & 1.536 & 1.134 & +0.402 \\
MiniMax-H3 Ref2VA~\cite{minimax2026h3} & 1--5 references, videos of up to 124 frames at about 768\x1344 & 20.675 & 2.039 & +18.636 \\
MiniMax-H3-Image~\cite{minimax2026h3image} & 1--3 reference images, short edge 1,024 pixels & 0.715 & 0.198 & +0.517 \\
Cosmos3-Nano~\cite{agarwal2026cosmos} & 1--4 controls, 81 frames, 480\x832 & 3.164 & 1.154 & +2.010 \\
\bottomrule
\end{tabular}
\end{table*}

\section{More quantitative results}
\label{sec:supp_more_quant}

This section extends Sec.~\ref{sec:exp_token_selection} and Tab.~\ref{tab:token_selection} of the main paper with additional quantitative results.

\subsection{Longer fine-tuning}
\label{sec:supp_long_ft}

\begin{table*}[!t]
\definecolor{sltlight}{RGB}{232,240,252}

\caption{\textbf{Longer fine-tuning.}
The SFT baselines of Tab.~\ref{tab:token_selection} are trained for 16K steps, while the Teacher and \method rows are identical to those in Tab.~\ref{tab:token_selection}.
The best results are \textbf{highlighted}, and the second best are \sd{underlined}.}
\label{tab:supp_long_ft}

\tablestyle{1pt}{1.05}
\newlength{\sltcol}\settowidth{\sltcol}{\scriptsize In/Outpaint}%
\newlength{\slttmp}\settowidth{\slttmp}{\scriptsize PCK@0.1\,$\uparrow$}\ifdim\slttmp>\sltcol\setlength{\sltcol}{\slttmp}\fi%
\newlength{\sltmeth}\settowidth{\sltmeth}{\footnotesize SFT, Reduced Res.}\addtolength{\sltmeth}{5pt}%
\newlength{\sltside}\setlength{\sltside}{10pt}%
\newlength{\sltprep}\settowidth{\sltprep}{\scriptsize Prep.}%
\settowidth{\slttmp}{\scriptsize (s)\,$\downarrow$}\ifdim\slttmp>\sltprep\setlength{\sltprep}{\slttmp}\fi%
\setlength{\tabcolsep}{\dimexpr(\textwidth-\sltmeth-\sltprep-12\sltcol-\sltside)/30\relax}%
\begin{tabular}{>{\centering\arraybackslash}p{\sltside} >{\raggedright\arraybackslash}p{\sltmeth} >{\centering\arraybackslash}p{\sltprep} *{12}{>{\centering\arraybackslash}p{\sltcol}}}
\toprule
& & & \multicolumn{3}{c}{Quality and Fidelity} & \multicolumn{9}{c}{Condition Following} \\
\cmidrule(lr){4-6}\cmidrule(l){7-15}
& \scriptsize Method & \scriptsize\makecell{Prep.\\(s)\,\dn} & \scriptsize FID\,\dn & \scriptsize FVD\,\dn & \scriptsize LPIPS\,\dn & \scriptsize \makecell{Identity\\DINOv3\,\up} & \scriptsize \makecell{Segm.\\mIoU\,\up} & \scriptsize \makecell{Depth\\si-RMSE\,\dn} & \scriptsize \makecell{Normal\\Angle\,\dn} & \scriptsize \makecell{Edges\\F1\,\up} & \scriptsize \makecell{In/Outpaint\\mPSNR\,\up} & \scriptsize \makecell{Pose\\PCK@0.1\,\up} & \scriptsize \makecell{Track\\TrajErr\,\dn} & \scriptsize \makecell{Grayscale\\$\Delta E_{00}$\,\dn} \\
\midrule
& Teacher & 0 & 22.4 & 118.0 & 0.273 & 0.288 & 0.888 & 0.647 & 4.72 & 0.445 & 20.93 & 0.912 & 13.54 & 6.63 \\
\midrule
& SFT, Random & $\sim$0 & 25.4 & 133.0 & 0.303 & 0.285 & 0.870 & 0.847 & 6.38 & 0.342 & 19.43 & \textbf{0.915} & 9.51 & 6.89 \\
& SFT, Reduced Res. & $\sim$0 & 34.1 & 213.7 & 0.424 & 0.284 & 0.673 & 1.494 & 13.40 & 0.124 & 14.81 & 0.846 & 13.97 & 11.81 \\
& SFT, DINO~\cite{Simoni2025DINOv3} & 23.8 & 26.3 & 132.0 & 0.330 & 0.285 & \sd{0.872} & 0.908 & 7.10 & 0.352 & 18.88 & 0.910 & \textbf{6.13} & 8.19 \\
& SFT, Saliency~\cite{parihar2026keep} & 75.0 & 27.5 & 138.8 & 0.360 & \textbf{0.291} & 0.836 & 1.267 & 10.22 & 0.315 & 16.65 & \sd{0.913} & 9.68 & 10.25 \\
\rowcolor{sltlight}\cellcolor{white}\multirow{-5}{*}{\rotatebox[origin=c]{90}{\scriptsize\textit{16K steps}}} & SFT, Codec (ours) & 1.13 & \textbf{22.9} & \sd{120.5} & \sd{0.281} & 0.287 & 0.822 & \sd{0.759} & \sd{5.47} & \sd{0.413} & \sd{20.22} & 0.905 & \sd{6.66} & \sd{6.78} \\
\midrule
\rowcolor{sltlight}\cellcolor{white}\rotatebox[origin=c]{90}{\scriptsize\textit{2K}} & \method (ours) & 1.13 & \sd{23.0} & \textbf{113.8} & \textbf{0.278} & \sd{0.289} & \textbf{0.883} & \textbf{0.706} & \textbf{5.44} & \textbf{0.417} & \textbf{20.43} & 0.911 & 7.80 & \textbf{6.53} \\
\bottomrule
\end{tabular}
\end{table*}

In Tab.~\ref{tab:supp_long_ft}, we train the SFT baselines of Tab.~\ref{tab:token_selection} for 16K steps instead of 2K to match the number of states that \method supervises in 2K steps with 8 states per rollout.
Note that this does not equalize the compute, since each \method step also runs the rollout and the teacher.
Even with this longer SFT training, \method achieves the best results in 8 of the 12 metrics.

\begin{figure*}[p]
\centering
\includegraphics[width=\textwidth,height=7.72in,keepaspectratio]{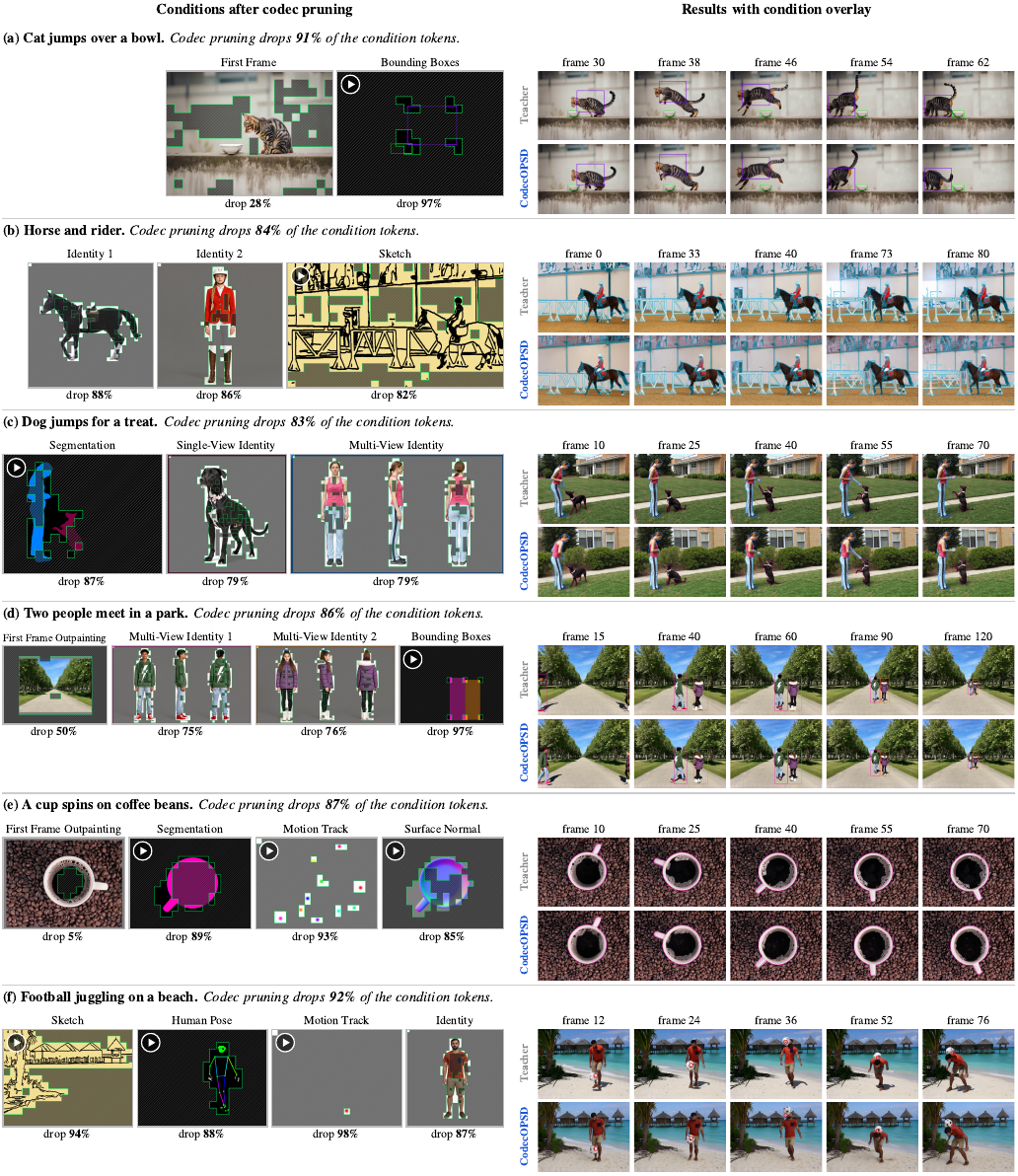}

\caption{\textbf{Qualitative results of \method on our multi-condition model.}
Each case shows the conditions with the tokens dropped by codec pruning darkened, followed by five frames of the teacher with all condition tokens and of \method with the remaining tokens.
A play icon marks a video condition shown by one frame, the results of (b) are cropped to 4:3, and all results are shown with condition overlay.
\method follows every condition as the teacher does while dropping 83--92\% of the condition tokens.}
\label{fig:supp_qual_ours}

\end{figure*}

\subsection{Condition tokens per condition}
\label{sec:supp_alloc}

\begin{table}[!tp]

\caption{\textbf{Condition tokens dropped by threshold-based pruning.}
We report the dropped tokens per condition on the test set of Tab.~\ref{tab:token_selection}.
Tokens denotes the number of tokens of one condition before pruning, where frame interpolation, continuation, and first-frame outpainting count the tokens of the given frames.
Blurring is not part of this test set and is evaluated in Fig.~\ref{fig:percond}.}
\label{tab:supp_alloc}

\tablestylesm{3.4pt}{1.05}
\begin{tabular}{l c c c c}
\toprule
Condition & Cases & Tokens & Dropped & Drop (\%) \\
\midrule
Motion track & 69 & 4,224 & 4,023 & 95 \\
Boxes & 66 & 4,224 & 3,906 & 92 \\
Identity & 69 & 1,484 & 1,280 & 86 \\
Segmentation & 64 & 4,224 & 3,593 & 85 \\
Human pose & 66 & 4,224 & 3,433 & 81 \\
Outpainting & 69 & 4,224 & 2,782 & 66 \\
Depth & 69 & 4,224 & 2,578 & 61 \\
Dense track & 64 & 4,224 & 2,483 & 59 \\
Sketch & 64 & 4,224 & 2,432 & 58 \\
Edges & 65 & 4,224 & 2,335 & 55 \\
First-frame outpainting & 68 & 384 & 201 & 52 \\
Inpainting & 68 & 4,224 & 2,110 & 50 \\
Normal & 68 & 4,224 & 2,029 & 48 \\
Grayscale & 63 & 4,224 & 1,739 & 41 \\
Interp. / continuation & 68 & 802 & 327 & 41 \\
\midrule
\textbf{All, per case} & \textbf{400} & \textbf{8,853} & \textbf{5,858} & \textbf{66} \\
\bottomrule
\end{tabular}
\end{table}

Threshold-based pruning has no fixed budget and drops 66\% of the condition tokens per case on average over the test set of Tab.~\ref{tab:token_selection}.
Random drop, DINO attention~\cite{Simoni2025DINOv3}, saliency~\cite{parihar2026keep,kummerer2025modeling}, and reduced resolution~\cite{tan2026ominicontrol2,zhang2025easycontrol,ben2026avcontrol}, the other pruning baselines in Tab.~\ref{tab:token_selection}, drop about the same share on average.
Among them, reduced resolution downsamples each condition and scales its 3D RoPE positions to the full grid.
As shown in Tab.~\ref{tab:supp_alloc}, codec pruning drops more tokens from sparse conditions and fewer from dense ones.
Therefore, the comparison measures both how tokens are allocated across conditions and how they are selected within each condition.

\subsection{Baseline timing}
\label{sec:supp_speed}

The selection times in Tabs.~\ref{tab:token_selection} and~\ref{tab:supp_prep} are measured on one A100 GPU with 6 CPU cores and 12 threads, which corresponds to the share of one GPU on an 8-GPU node.
Each case is timed twice, and we report the faster run.
DINO attention and saliency run the 7B DINOv3~\cite{Simoni2025DINOv3} and DeepGaze~\cite{kummerer2025modeling}, respectively, on the GPU without further speed optimization.
As shown in Tab.~\ref{tab:supp_prep}, codec pruning runs on the CPU in parallel with the VAE and text encoding on the GPU and finishes first on average for all four base models.

\section{Qualitative results}
\label{sec:supp_qual}

In Fig.~\ref{fig:supp_qual_ours}, we provide additional results of our multi-condition model on condition combinations not shown in the main paper.
In all cases, \method follows each condition as the teacher does while dropping 83--92\% of the condition tokens, and codec pruning retains the tokens that carry the content of each condition.
Codec pruning keeps the corners of each moving box in (a) and (d) and the subject of each identity reference in (b), (c), (d), and (f), while it drops the plain background and the empty part of the outpainting canvas in (d).
Over time, the remaining tokens follow the changes in each condition, so the blank frames between the sketch drawings in (b) and the repeated frames of the held sketch in (f) are dropped.
For more results, we recommend that readers visit our project page, which presents these cases as videos.

\section{More ablation studies}
\label{sec:supp_ablation_more}

This section provides more details on the teacher-student gap ablation in Sec.~\ref{sec:exp_gap} of the main paper, followed by two more ablation studies.
For the larger gap in Tab.~\ref{tab:gap}, the student drops 90\% of the tokens of every condition during training, and for the smaller gap, the teacher is also pruned with half of the student's codec thresholds.
At test time, both variants and \method use the same threshold-based pruning as in Tab.~\ref{tab:token_selection}, with the thresholds in Tab.~\ref{tab:supp_codec_settings}, so they receive identical inputs and differ in training alone.
We then study a more privileged teacher in Supp.~\ref{sec:supp_privileged} and the number of supervised states in Supp.~\ref{sec:supp_onestate}.

\subsection{Privileged teacher}
\label{sec:supp_privileged}

\begin{table}[t]

\caption{\textbf{A more privileged teacher on the distilled version of our internal model.}
Both students are trained for 2K steps and tested on the 200 test clips of Tab.~\ref{tab:token_selection} with one or two conditions.
Each ratio to GT divides the value of the generated clip by that of the ground truth, and we report its median over clips for contrast and saturation and its mean for Laplacian sharpness.
The better results are highlighted.}
\label{tab:supp_privileged_distilled}

\tablestylesm{2pt}{1.05}
\resizebox{\columnwidth}{!}{%
\begin{tabular}{@{}l ccc ccc@{}}
\toprule
& \multicolumn{3}{c}{Quality} & \multicolumn{3}{c}{Ratio to GT ($\to$1)} \\
\cmidrule(lr){2-4}\cmidrule(l){5-7}
& FID\,\dn & FVD\,\dn & Aesthetic\,\up & Contrast & Saturation & Sharpness \\
\midrule
\cellcolor{ourgray}\textbf{\method} & \cellcolor{ourgray}\textbf{43.1} & \cellcolor{ourgray}\textbf{307.7} & \cellcolor{ourgray}\textbf{0.530} & \cellcolor{ourgray}\textbf{1.02} & \cellcolor{ourgray}\textbf{1.03} & \cellcolor{ourgray}\textbf{1.78} \\
\makecell[l]{\method with a more\\ \hspace{0.5em}privileged teacher} & 57.9 & 370.2 & 0.502 & 0.96 & 0.89 & 2.25 \\
\bottomrule
\end{tabular}}
\end{table}

\begin{figure*}[!t]
\centering
\includegraphics[width=0.84\textwidth]{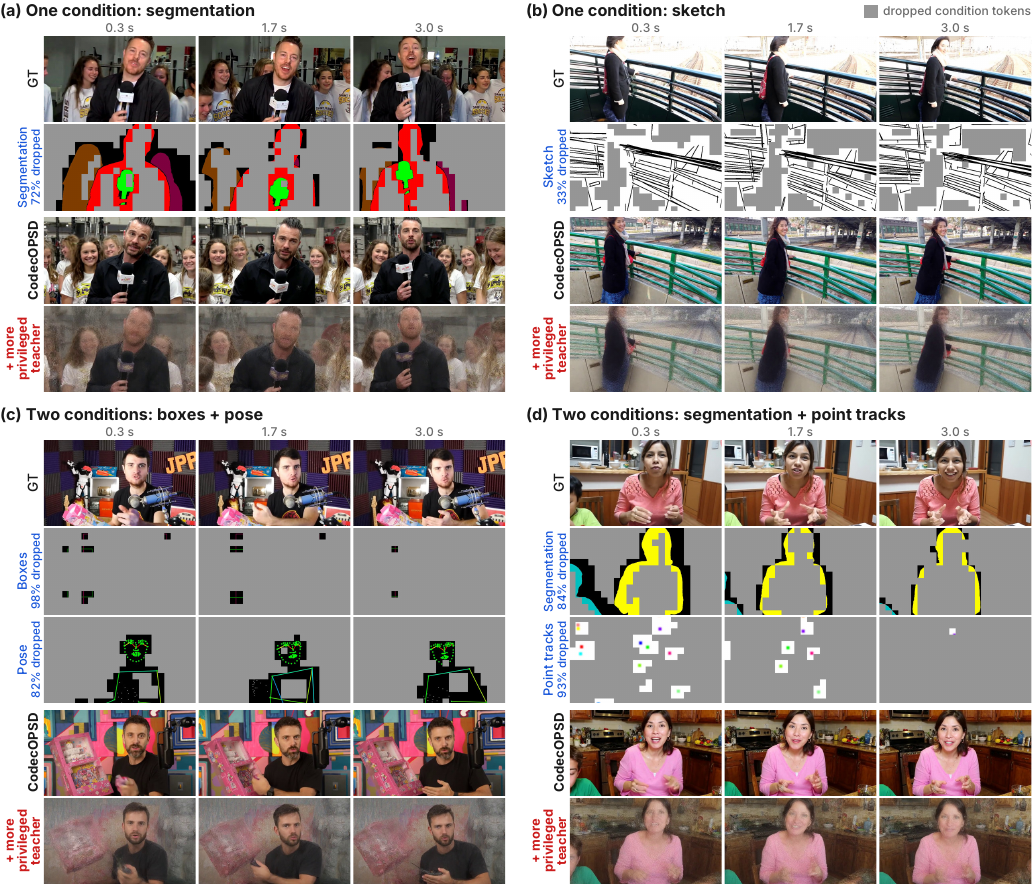}

\caption{\textbf{A more privileged teacher degrades the student on cases with few conditions.}
Each panel shows the ground truth, each condition with its drop rate and with the tokens dropped by threshold-based pruning in grey, and the results of \method and of its variant with a more privileged teacher, which also sees the other conditions of the clip.
Both students are trained for 2K steps on the distilled version of our internal model and receive the same input and remaining condition tokens, so they differ in the teacher used for training alone.}
\label{fig:supp_privileged}

\end{figure*}

\begin{table}[t]

\caption{\textbf{Eight versus one supervised state per rollout on our multi-condition model.}
After 2K training steps, eight states of the 30-step grid and one state of a 16-step grid achieve similar results, except on track following, while one state makes each training step 3.1\x\ faster on 16 GPUs of the same type.
The better metric results are highlighted.}
\label{tab:supp_onestate}

\tablestyle{4pt}{1.05}
\begin{tabular}{@{}l l >{\columncolor{ourgray}}c c@{}}
\toprule
& & 8 states (\method) & 1 state \\
\midrule
\multicolumn{2}{@{}l}{Training time (s / step)} & 34.6 & 11.0 \\
\midrule
\multirow{3}{*}{\makecell[l]{Quality and\\Fidelity}} & FID\,\dn & 23.0 & \textbf{22.9} \\
& FVD\,\dn & \textbf{113.8} & 115.4 \\
& LPIPS\,\dn & \textbf{0.278} & 0.282 \\
\midrule
\multirow{6}{*}{\makecell[l]{Condition\\Following}} & DINOv3\,\up & \textbf{0.289} & \textbf{0.289} \\
& si-RMSE\,\dn & 0.706 & \textbf{0.691} \\
& Edges F1\,\up & \textbf{0.417} & \textbf{0.417} \\
& mPSNR\,\up & 20.43 & \textbf{20.50} \\
& PCK\,\up & \textbf{0.911} & 0.910 \\
& TrajErr\,\dn & \textbf{7.80} & 9.15 \\
\bottomrule
\end{tabular}
\end{table}

In \method, the teacher sees all tokens of the conditions given to the student.
A more privileged teacher additionally sees the other conditions of the clip, with up to five conditions in total, including pixel conditions such as inpainting or the first frame.
We denote the tokens of these additional conditions, which the student does not receive, as $\mathcal C^{+}$, and the teacher in Eq.~\ref{eq:teacher_student} can thus be formulated as:
\begin{equation}
  v_\mathrm{T}^{+}(x,\sigma)=v_\theta\big(x,\sigma;\tilde{\mathcal C}\cup\mathcal C^{\star}\cup\mathcal C^{+}\big),
  \label{eq:privileged}
\end{equation}
while the student, the rollout in Eq.~\ref{eq:rollout}, and the loss in Eq.~\ref{eq:opsd} remain unchanged.
For this study, we train both this variant and \method on the distilled version of our internal model described in Supp.~\ref{sec:supp_distilled}.
As shown in Fig.~\ref{fig:supp_privileged}, the more privileged teacher degrades the student outputs on test clips with one or two conditions, leading to washed-out colours and degraded details.
Tab.~\ref{tab:supp_privileged_distilled} confirms this observation, where the FID increases from 43.1 to 57.9 and the saturation decreases to 0.89 of the ground truth.

Therefore, we do not adopt a more privileged teacher.
Besides this degradation, its design is also hard to define.
For our multi-condition model, the type and number of extra conditions for the teacher are already free design choices.
For a model whose task changes with every request, such as MiniMax-H3 Ref2VA~\cite{minimax2026h3}, there is no natural set of extra conditions to add.
In contrast, \method provides the teacher with the same conditions as the student and thus avoids this choice.

\subsection{One versus eight supervised states}
\label{sec:supp_onestate}

Both settings share the rollout in Eq.~\ref{eq:rollout} and the per-state loss $\ell_k$ in Eq.~\ref{eq:opsd}, and differ in the set $\mathcal N$ of supervised states, which can be formulated as:
\begin{equation}
  \mathcal N_{\mathrm{dense}}=\{1,\dots,N\},\qquad\mathcal N_{\mathrm{single}}=\{k\},\ \ k\sim\mathcal U\{1,\dots,\tilde K\},
  \label{eq:onestate_sets}
\end{equation}
where the dense setting of the main paper supervises the first $N=8$ states of the $K=30$ states of the inference grid, all at high noise, and the single-state setting draws one state from a grid of $\tilde K=16$ noise levels.
Accordingly, the objective in Eq.~\ref{eq:opsd} becomes:
\begin{equation}
  \begin{aligned}
  \mathcal L_{\mathrm{dense}}(\phi)&=\mathbb E_{\mathcal C,\epsilon}\Big[\frac{1}{N}\sum_{k=1}^{N}\ell_k\Big],\\
  \mathcal L_{\mathrm{single}}(\phi)&=\mathbb E_{\mathcal C,\epsilon}\,\mathbb E_{k\sim\mathcal U\{1,\dots,\tilde K\}}\big[\ell_k\big],
  \end{aligned}
  \label{eq:onestate}
\end{equation}
where $\ell_k=\|v_\mathrm{S}(x_k,\sigma_k)-v_\mathrm{T}(x_k,\sigma_k)\|_2^2$ compares the student and the teacher at the $k$-th state of the rollout of the student.
In the single-state setting, every state of the $\tilde K$-step grid is thus supervised over the course of training.
For each sample, the dense setting rolls out to $x_N$ and performs $N$ teacher passes and $N$ student passes with gradients, while the single-state setting rolls out to $x_k$ and performs one of each.
Therefore, the single-state setting makes each training step roughly 3.1\x\ faster.
DanceOPD~\cite{zhou2026danceopd} also supervises a single state per rollout, but samples its noise level from a distribution biased toward low noise.

As shown in Tab.~\ref{tab:supp_onestate}, the single-state setting performs close to the dense setting on most metrics of our multi-condition model and is worse on track following, with a TrajErr of 9.15 against 7.80.
Therefore, the single-state setting is a sound choice when training compute is limited.

The two settings thus differ little in quality, while the single-state setting saves about two thirds of the training time per step.
This saving matters much more on MiniMax-H3 Ref2VA~\cite{minimax2026h3}, whose DiT has 33B parameters, about twice the 15B of our internal model, and whose inputs add long reference videos and a 32B VLM.
On 16 H200 GPUs, one of its training steps takes about 39\,s with a single supervised state and about 132\,s with eight states.
Therefore, the \method LoRAs of MiniMax-H3 Ref2VA and MiniMax-H3-Image~\cite{minimax2026h3image} adopt the single-state setting, which draws the supervised state at random from a 16-step grid and from the released 6-step grid, respectively, so that every noise level is supervised over the course of training.

\begin{table*}[!t]

\caption{\textbf{Training tasks of the \method LoRA for MiniMax-H3 Ref2VA~\cite{minimax2026h3}.}
(a) lists the 41 training tasks in three groups, which edit, transfer, or restore an input clip, generate from reference images and storyboards, or follow control inputs.
Each task lists its input images (I) and videos (V) and its share of the training steps, and \dag{} marks the 24 tasks of the earlier LoRA in Supp.~\ref{sec:supp_addon}.
(b) lists the output resolutions, and (c) lists the frames, image sizes, and inputs per sample.}
\label{tab:supp_tasks_minimax}

\tablestylesm{2.6pt}{1.0}
\textbf{(a) Tasks}\par\smallskip
\begin{tabular}{l >{\raggedright\arraybackslash}p{0.225\textwidth} c r @{\hspace{8pt}} >{\raggedright\arraybackslash}p{0.148\textwidth} >{\raggedright\arraybackslash}p{0.212\textwidth} c r}
\toprule
Task & Inputs $\to$ output & Refs & \% & Task & Inputs $\to$ output & Refs & \% \\
\midrule
\multicolumn{4}{l}{\textbf{Edit, transfer, and scene (6)}} & \multicolumn{4}{l}{\textbf{Reference images and storyboards (cont.)}} \\
Edit\dag & clip + edit instruction $\to$ edited clip & 1V & 19.2 & Composition, 4 refs & person + 2 products + scene $\to$ video & 4I & 0.7 \\
Style\dag & clip $\to$ same clip in a new style & 1V & 2.0 & Storyboard sheet & numbered sheet $\to$ panels in order & 1I & 0.6 \\
Motion transfer\dag & clip $\to$ new subject, same motion & 1V & 8.5 & Storyboard video & storyboard video $\to$ live action & 1V & 1.2 \\
Camera transfer\dag & clip $\to$ new place, same camera move & 1V & 5.1 & Coarse line art & line-art video $\to$ live action & 1V & 1.1 \\
Scene\dag & subject + scene image $\to$ subject there & 2I & 6.4 & Storyboard + 1 person & front image + storyboard $\to$ video & 1I1V & 0.8 \\
Restoration\dag & blurred or low-res clip $\to$ clean clip & 1V & 4.5 & Storyboard + 2 people & two fronts + storyboard $\to$ video & 2I1V & 0.2 \\
\cmidrule(r){1-4}
\multicolumn{4}{l}{\textbf{Reference images and storyboards (17)}} & Multi-shot story & 2--3-shot clip $\to$ new cast, same cuts & 1V & 4.0 \\
\cmidrule(l){5-8}
Subject + camera & subject image + camera clip $\to$ video & 1I1V & 3.2 & \multicolumn{4}{l}{\textbf{Control (18)}} \\
Subject + action & subject image + action clip $\to$ video & 1I1V & 2.3 & First / last frame\dag & given frames $\to$ full clip & 1V & 6.8 \\
Subject edit & clip + subject $\to$ subject swapped or added & 1I1V & 4.2 & Identity\dag & 1--4 subject images $\to$ video & 1--4I & 4.2 \\
Style image & style image + prompt $\to$ styled video & 1I & 4.8 & Depth\dag, sketch\dag, pose\dag & condition video $\to$ video & 1V & 1.3 each \\
Outfit & person image + new outfit in text $\to$ video & 1I & 1.1 & Normal\dag, edge\dag, dense track\dag & condition video $\to$ video & 1V & 0.8 each \\
Character sheet & multi-view sheet $\to$ video of the character & 1I & 1.1 & Grayscale\dag, segmentation\dag & condition video $\to$ video & 1V & 0.8 each \\
Grounding & group picture + named person $\to$ video & 1I & 1.1 & Inpaint\dag & masked clip $\to$ area filled & 1V & 1.1 \\
Object & product image $\to$ object or pattern in video & 1I & 1.0 & Identity + first / last frame\dag & images + given frames $\to$ video & 1--4I1V & 1.5 \\
Composition & person + product + scene $\to$ video & 3I & 1.4 & Identity + pose / depth / sketch\dag & images + condition $\to$ video & 1--4I1V & 0.6 each \\
Grounded composition & group frame + product + scene $\to$ video & 3I & 0.4 & First / last frame + pose / depth / sketch\dag & frames + condition $\to$ video & 2V & 0.6 each \\
\bottomrule
\end{tabular}

\medskip
\begin{minipage}[t]{0.4\textwidth}
\centering\textbf{(b) Resolution}\par\smallskip
\begin{tabular}{l c c}
\toprule
Output & Steps (\%) & Smaller conditions \\
\midrule
480\x640 & 64 & 384\x512 \\
480\x832 & 25 & 384\x672 \\
704\x384 & 5 & 576\x320 \\
384\x832 & 1 & 320\x672 \\
768\x1344 & 5 & 608\x1088 \\
\bottomrule
\end{tabular}
\end{minipage}\hfill
\begin{minipage}[t]{0.56\textwidth}
\centering\textbf{(c) Frames, images, and inputs per sample}\par\smallskip
\begin{tabular}{l l}
\toprule
Frames & 56 on 70\% of the steps, 124 on 30\% \\
Images & aspect ratio kept, long side 512--2048\,px \\
Inputs & videos alone 63\%, images alone 23\%, images and a video 14\% \\
\bottomrule
\end{tabular}
\end{minipage}
\end{table*}

\section{MiniMax-H3 Ref2VA}
\label{sec:supp_minimax}
\label{sec:supp_train_other}

This section extends Sec.~\ref{sec:exp_minimax} of the main paper with the training details, a comparison with Training-Free Codec, and the use of add-on LoRAs and few-step models.

\subsection{Training details}
\label{sec:supp_train_details}

Since the teacher supervises the student on the student's own rollout, each training sample consists of the inputs and a prompt, and no target output is required.
The \method LoRA has rank 128, about 655M parameters, and is applied to the condition tokens.
It is trained on the 41 tasks in Tab.~\ref{tab:supp_tasks_minimax} for 3,000 steps with a learning rate of $5\times10^{-5}$, where each step supervises one state of a 16-step grid.
Due to GPU memory limits, each node of 8 GPUs dedicates one GPU to Qwen3-VL-32B~\cite{bai2025qwen3}, the VLM of MiniMax-H3 Ref2VA, and trains on the other seven, resulting in a batch size of 14 on 16 GPUs.
To save compute, 70\% of the training steps generate 56 frames, while the others generate 124 frames.
The student receives the reference tokens that remain after threshold-based pruning, and its Qwen3-VL drops reference vision tokens with the rule of Supp.~\ref{sec:supp_eval_minimax}.
To make the student more robust, part of the training steps further drop some of the remaining reference video tokens through priority-based sorting.
An earlier LoRA, which is used in Supp.~\ref{sec:supp_addon}, is trained on the 24 tasks marked \dag{}.

\subsection{Training-Free Codec versus \method}
\label{sec:supp_trainfree}

\begin{figure*}[p]
\centering
\includegraphics[width=\textwidth,height=7.85in,keepaspectratio]{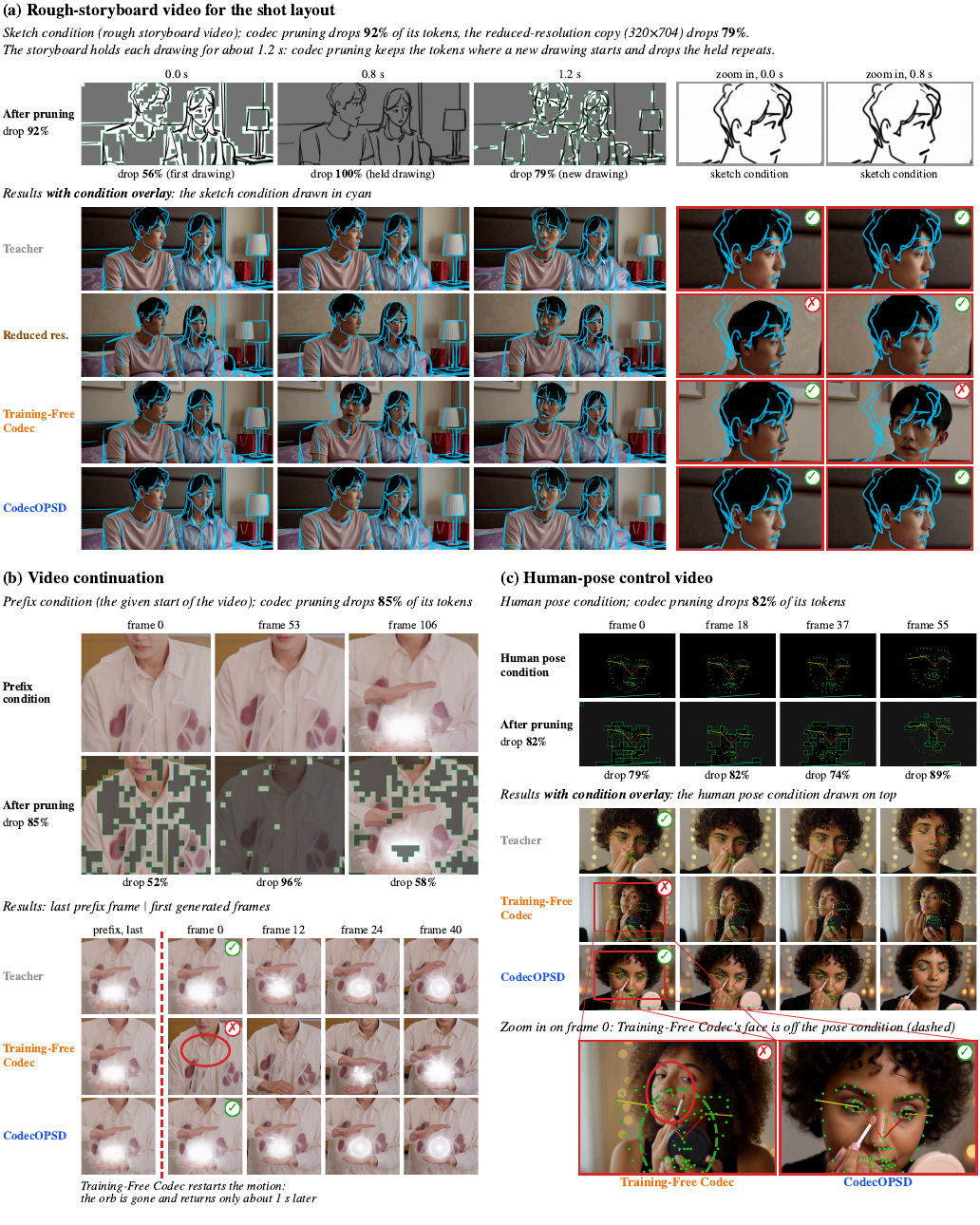}

\caption{\textbf{Training-Free Codec and \method on MiniMax-H3 Ref2VA.}
We show (a) a rough-storyboard video, (b) a video continuation, and (c) a human-pose control video, with condition overlay in (a) and (c).
Both models receive the same remaining reference tokens, inputs, and seed.}
\label{fig:supp_trainfree_minimax}

\end{figure*}

Codec pruning can be applied to a released model without training, which we refer to as Training-Free Codec.
In this setting, the released weights receive the remaining condition tokens alone.
Since the released model is not trained on references with missing tokens, it cannot tell what the dropped tokens represent.
As shown in Fig.~\ref{fig:supp_trainfree_minimax}, with 82--92\% of the reference tokens dropped on MiniMax-H3 Ref2VA~\cite{minimax2026h3}, Training-Free Codec turns a character toward the camera against the storyboard in (a), restarts a motion that has already begun in the given clip in (b), and moves the face away from the human pose condition in (c).
In contrast, \method stays close to the teacher with the same remaining tokens.
Reduced resolution drops fewer reference tokens, 79\% compared with 92\%, but still shifts the whole shot in (a).
These failures motivate the need for training.

\subsection{Add-on LoRAs and few-step models}
\label{sec:supp_addon}

\begin{figure*}[p]
\centering
\makebox[\textwidth][c]{\begin{minipage}{7.5in}
\includegraphics[width=\linewidth]{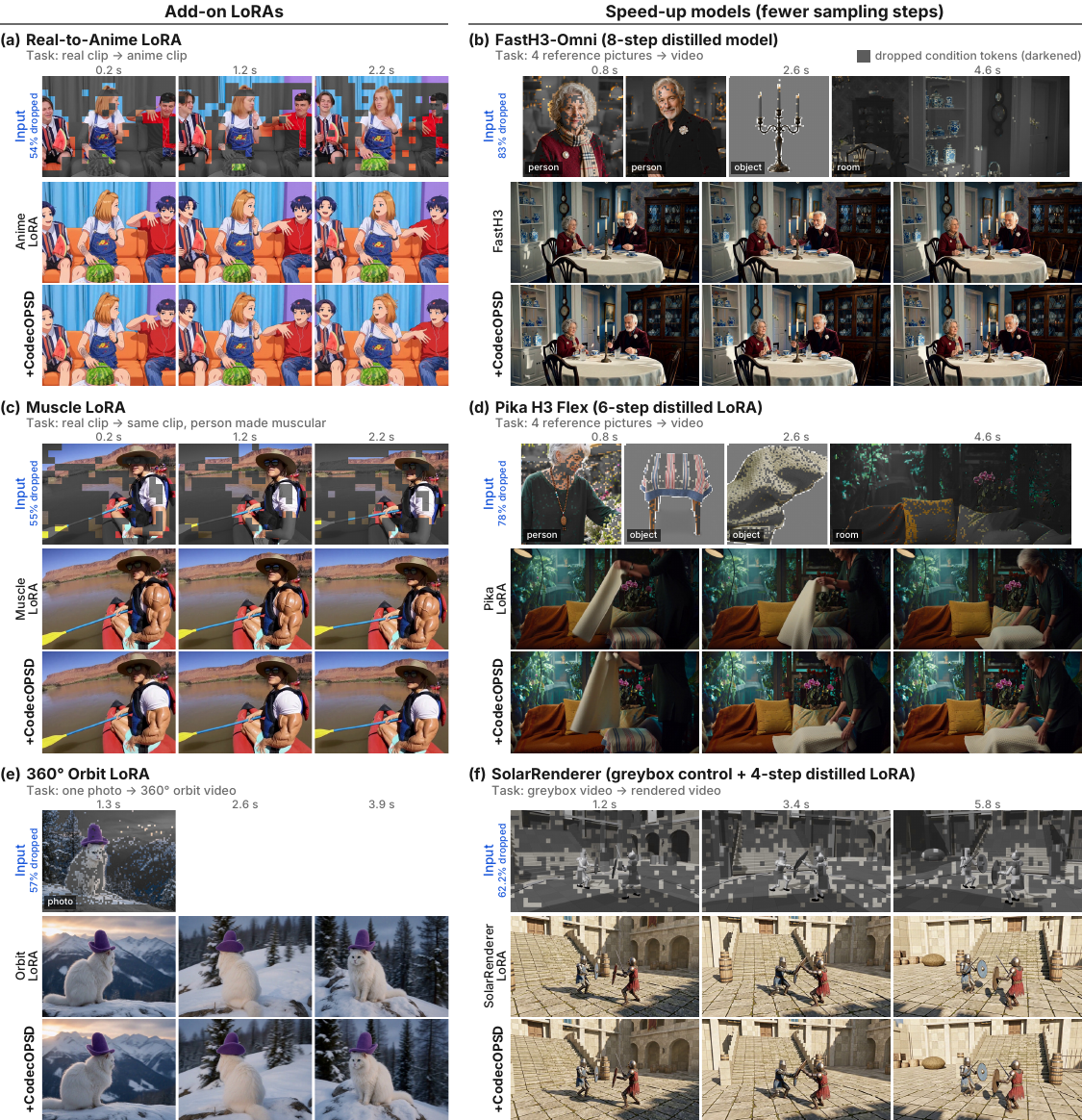}

\caption{\textbf{\method with add-on LoRAs and few-step models on MiniMax-H3 Ref2VA, without further training.}
Each panel shows the input, where the darkened regions mark the condition tokens dropped by codec pruning, followed by the results of the add-on LoRA or the few-step model with all condition tokens and with \method, generated from the same input and seed.
Panels (a), (c), and (e) apply the Real-to-Anime, Muscle, and 360\textdegree{} Orbit LoRAs, and panels (b), (d), and (f) use FastH3-Omni, Pika H3 Flex, and SolarRenderer.
In every panel, \method keeps the effect of the add-on LoRA or the few-step model with most condition tokens dropped.}

\label{fig:supp_addon_minimax}
\end{minipage}}
\end{figure*}

More visualizations are available on our project page.
Released models are often used with third-party add-on LoRAs or as faster few-step versions.
We apply codec pruning and our \method LoRA to both settings without further training, with add-on LoRAs in Fig.~\ref{fig:supp_addon_minimax}a, c, and e and few-step models in Fig.~\ref{fig:supp_addon_minimax}b, d, and f.
The add-on LoRA acts on all tokens as it was trained to, while our LoRA and threshold-based pruning act on the condition tokens.

\textbf{Add-on LoRAs.}
On MiniMax-H3 Ref2VA~\cite{minimax2026h3}, the community Real-to-Anime\footnote{\url{https://huggingface.co/akhaliq/Minimax-H3-ref2v_Real_2_Anime}}, Muscle\footnote{\url{https://huggingface.co/ostris/minimax_h3_ref2va_jacked_lora}}, and 360\textdegree{} Orbit\footnote{\url{https://huggingface.co/MATLOWAI/MiniMax-H3-ORB360-CardSpin}} LoRAs retain their effects with 54--57\% of the condition tokens dropped, as seen in the anime look, the muscles, and the full orbit back to the starting view.
Compared with the add-on results with all condition tokens and the same input and seed, the median PSNR is 20.3, 25.9, and 18.3\,dB over 9, 5, and 4 clips, respectively, while MiniMax-H3 Ref2VA without the add-on achieves 15.7, 9.1, and 9.7\,dB.

\textbf{Few-step models.}
Our MiniMax-H3 Ref2VA LoRA also works unchanged on few-step models trained by others, as in Fig.~\ref{fig:scaling} top.
Fig.~\ref{fig:supp_addon_minimax} shows it with the 8-step FastH3-Omni of FastVideo~\cite{pmlr-v267-zhang25m}\footnote{\url{https://huggingface.co/FastVideo/FastVideo-FastH3-Omni-8-Step-V1}}, the 6-step Pika H3 Flex\footnote{\url{https://huggingface.co/Pika-Labs/Pika-H3-Flex}}, and the 4-step SolarRenderer~\cite{fang2026solarrenderer} with greybox control.
The two speedups multiply, since few-step models reduce the denoising steps, while codec pruning reduces the condition tokens at each step.
On these cases at 768\x1344, codec pruning makes the denoising of FastH3-Omni, Pika H3 Flex, and SolarRenderer another 1.50\x, 1.46\x, and 1.79\x\ faster, or 9.19\x, 11.20\x, and about 21.9\x\ faster than the released 50-step model.

\begin{table*}[!tp]

\caption{\textbf{Training data of the \method LoRA for MiniMax-H3-Image~\cite{minimax2026h3image}.}
(a) lists the 22,700 training samples by task, grouped into edits of one image, composition from 2 to 5 images, and group photos with 6 to 16 images.
Samples marked \dag{} are used in steps 1,501 to 2,400, samples marked \ddag{} in steps 2,401 to 3,000, and the other samples in all 3,000 steps.
(b) lists the output sizes and the short sides of the reference images, and (c) lists the share of training steps by the number of reference images in each stage.}

\label{tab:supp_tasks_h3img}

\tablestylesm{2.6pt}{1.0}
\textbf{(a) Tasks}\par\smallskip
\begin{tabular}{>{\raggedright\arraybackslash}p{0.17\textwidth} >{\raggedright\arraybackslash}p{0.19\textwidth} c r @{\hspace{8pt}} >{\raggedright\arraybackslash}p{0.115\textwidth} >{\raggedright\arraybackslash}p{0.245\textwidth} c r}
\toprule
Task & Inputs $\to$ output & Refs & Samples & Task & Inputs $\to$ output & Refs & Samples \\
\midrule
\multicolumn{4}{l}{\textbf{Edits of one image (26)}} & \multicolumn{4}{l}{\textbf{Composition from 2 to 5 images (cont.)}} \\
Art style, line art, sketch, greybox, edges, turnaround & image $\to$ drawing, render, edges, or views, and edges $\to$ photo & 1 & 317 & Product, scene, and style & products or people, optional scene, + style image $\to$ image in that style & 2--5 & 1,581 \\
Object add, remove, replace, restyle, circle edit & image $\to$ one named or circled object changed & 1 & 281 & Product in a new scene & 1--2 products, optional person, + scene $\to$ product photo & 2--3 & 1,400 \\
Brighten, relight, weather, colourize, deblur & image $\to$ new light, weather, or colour, or sharp & 1 & 280 & Outfit & person + clothes from 1--2 images, optional accessory $\to$ dressed person & 2--4 & 1,151 \\
Inpaint, outpaint, zoom in, new background & image $\to$ filled, extended, zoomed in, or new background & 1 & 216 & Four or five subjects & 4--5 people, animals, or objects, optional scene $\to$ one image & 4--5 & 1,064 \\
Expression, garment extraction & person $\to$ new expression, or the garment alone & 1 & 112 & Style reference & image + style image $\to$ same image in that style & 2 & 299 \\
Scene text replace, remove & image $\to$ text replaced or removed & 1 & 85 & Colour-grade transfer & image + graded image $\to$ same image in that grade & 2 & 150 \\
\cmidrule(l){5-8}
Next shot, 2\x2 storyboard & frame $\to$ next shot or four panels & 1 & 109 & \multicolumn{4}{l}{\textbf{Group photos from 6 to 16 images (3)}} \\
\cmidrule(r){1-4}
\multicolumn{4}{l}{\textbf{Composition from 2 to 5 images (8)}} & Group photo\dag & 5--9 people, optional scene $\to$ group photo & 6--9 & 5,600 \\
Subjects into a scene & 1--3 people, animals, or objects, optional scene $\to$ one image & 2--4 & 4,150 & Group photo\ddag & 5--9 people, optional scene $\to$ group photo & 6--9 & 1,200 \\
Group photo & 2--5 people, optional scene $\to$ group photo & 2--5 & 2,805 & Group photo with objects\ddag & people + 1--2 objects, optional scene $\to$ one image & 10--16 & 1,900 \\
\bottomrule
\end{tabular}

\medskip
\begin{minipage}[t]{0.4\textwidth}
\centering\textbf{(b) Resolution}\par\smallskip
\begin{tabular}{l r}
\toprule
Output & Samples \\
\midrule
1536\x2720, 1664\x2496, 1760\x2368 & 17,586 \\
2048\x2048 & 1,693 \\
2368\x1760, 2496\x1664, 2720\x1536 & 3,421 \\
\midrule
Reference short side & Images \\
\midrule
about 1024 & 90,781 \\
about 1536 & 18,247 \\
about 2048 & 8,612 \\
\bottomrule
\end{tabular}
\end{minipage}\hfill
\begin{minipage}[t]{0.56\textwidth}
\centering\textbf{(c) Share of training steps (\%)}\par\smallskip
\begin{tabular}{l c c c c}
\toprule
& \multicolumn{4}{c}{References per sample} \\
\cmidrule(l){2-5}
Steps & 1 & 2--5 & 6--9 & 10--16 \\
\midrule
1--1,500 & 10 & 90 & 0 & 0 \\
1,501--2,400 & 8 & 72 & 20 & 0 \\
2,401--3,000 & 4.5 & 36 & 10.5 & 49 \\
All 3,000 & 8.3 & 73.8 & 8.1 & 9.8 \\
\bottomrule
\end{tabular}
\end{minipage}
\end{table*}

\section{MiniMax-H3-Image}
\label{sec:supp_h3img}

This section extends Sec.~\ref{sec:exp_h3image} of the main paper, with the same structure as the previous section.

\subsection{Training details}
\label{sec:supp_h3img_train}

Since MiniMax-H3-Image is released as a 6-step distilled model, the student rolls out this 6-step sampler, and each training step supervises one of its 6 states, drawn at random.
The \method LoRA has rank 128 on the attention and MLP projections, about 655M parameters, and is applied to the reference tokens.
It is trained from the released model for 3,000 steps on samples with 1 to 16 reference images, using a batch size of 14 and a learning rate of $5\times10^{-5}$.
Tab.~\ref{tab:supp_tasks_h3img} lists these samples, whose number of reference images grows over three training stages.
The student receives the reference tokens that remain after threshold-based pruning in the DiT, while its Qwen3-VL-32B~\cite{bai2025qwen3} also drops reference vision tokens.

\subsection{Training-Free Codec versus \method}
\label{sec:supp_trainfree_h3img}

\begin{figure*}[p]
\vspace{16pt}%
\centering
\includegraphics[width=\textwidth]{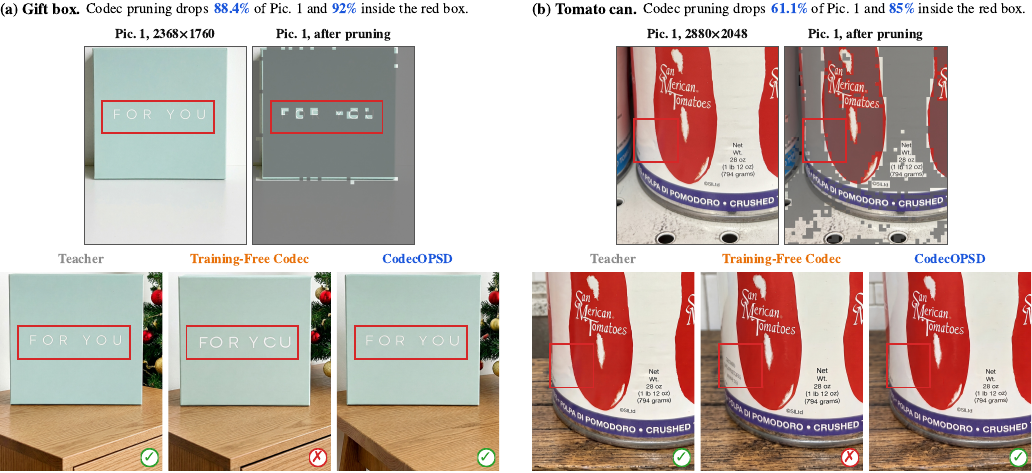}

\caption{\textbf{Training-Free Codec and \method on MiniMax-H3-Image~\cite{minimax2026h3image}.}
Both models receive the same remaining tokens after codec pruning, as well as the same inputs and seed, and all images show the same crop.
Within the red box, Training-Free Codec redraws the dropped thin lettering and generates garbled print on the blank label, while \method matches the teacher.}

\label{fig:supp_trainfree_h3img}
\end{figure*}

\begin{figure*}[t]
\centering
\includegraphics[width=0.76\textwidth]{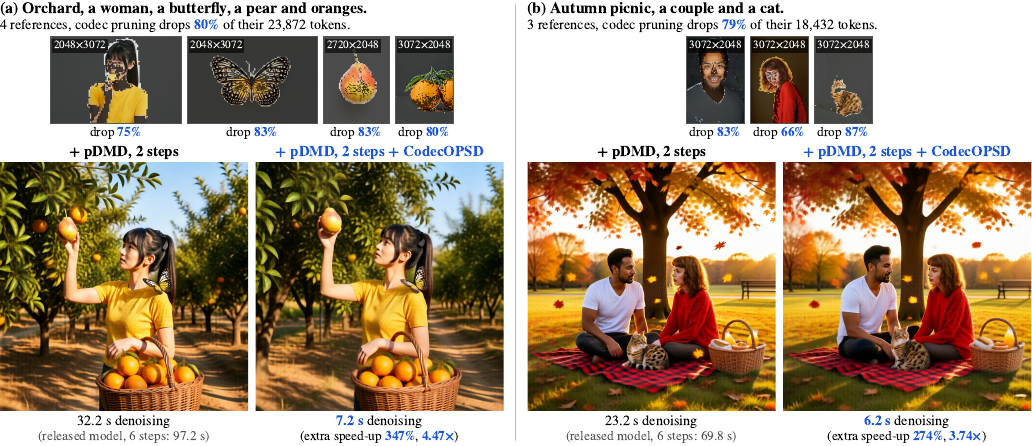}

\caption{\textbf{Stacked speedups on MiniMax-H3-Image LoRA without further training.}
Each case shows the reference images at a short side of 2,048 pixels with the tokens dropped by codec pruning darkened, followed by the results of 2-step PDMD~\cite{wang2026pdmd} alone and with \method, generated from the same inputs and seed.
Times are denoising seconds per image, and each label shows the extra speedup of \method over PDMD alone, which is large here because the references are at a high resolution.}

\label{fig:supp_addon_h3img}
\end{figure*}

On MiniMax-H3-Image~\cite{minimax2026h3image}, the errors of Training-Free Codec occur either inside the dropped regions or in the framing of the image.
As shown in Fig.~\ref{fig:supp_trainfree_h3img}~(a), codec pruning drops 88.4\% of the tokens of the gift box image, including its thin white lettering, and Training-Free Codec renders the lettering larger and outlined.
In (b), codec pruning drops 61.1\% of the tokens of the can photo, including a blank part of the label, where Training-Free Codec generates garbled print.
Since the released model has not learned to read images with dropped tokens, it cannot tell that a dropped region of a label is flat and fills it with invented letters.
In contrast, the teacher and \method preserve both the lettering and the label as in the given images.
In other cases, Training-Free Codec changes the framing of the edited image by zooming in and cropping part of the scene.
As with MiniMax-H3 Ref2VA, these failures motivate the need for training.

\subsection{Few-step model}
\label{sec:supp_h3img_fewstep}

For MiniMax-H3-Image~\cite{minimax2026h3image}, the few-step model we found in the community is the 2-step PDMD LoRA\footnote{\url{https://huggingface.co/akhaliq/minimax-h3-image-pdmd-2step}}, a repackaged version of the LoRA that PDMD~\cite{wang2026pdmd} distilled from MiniMax-H3.\footnote{\url{https://huggingface.co/pdmd2026/pdmd_2NFE_lora}}
With our LoRA added without further training, every reference in Fig.~\ref{fig:supp_addon_h3img} is reproduced.
With references at a short side of 2,048 pixels, codec pruning makes the denoising of PDMD another 4.47\x\ and 3.74\x\ faster, or 13.5\x\ and 11.3\x\ faster than the released 6-step model.
End to end, PDMD with \method is 1.79\x\ and 1.54\x\ faster than PDMD alone.
The waxy surfaces originate from PDMD and appear the same with and without \method.

\section{Cosmos3-Nano}
\label{sec:supp_cosmos}

This section extends Sec.~\ref{sec:exp_cosmos} of the main paper with the training details and qualitative results.

\subsection{Training details}
\label{sec:supp_cosmos_train}

\begin{figure*}[t]
\centering
\includegraphics[width=0.94\textwidth]{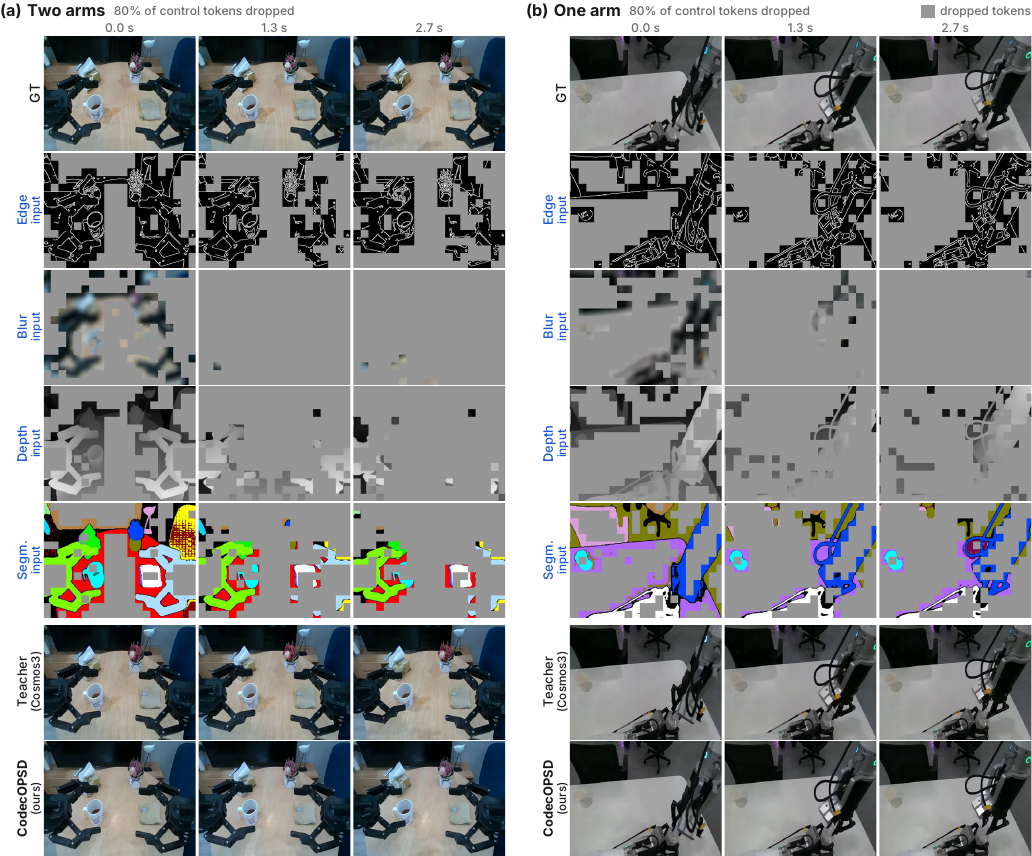}

\caption{\textbf{Codec pruning for Cosmos3-Nano transfer on robot clips with all four controls together.}
The rows show the ground truth, the four controls with the dropped tokens in grey, the teacher, which is Cosmos3-Nano~\cite{agarwal2026cosmos} with all tokens, and \method, where codec pruning drops 80\% of the control tokens in both clips.
The PAI-Bench-C~\cite{zhou2026pai} robot clips have 81 frames at 480\x832, are cropped to 4:3, and use the same input and seed for both models.
This setting lies outside Tab.~\ref{tab:cosmos}, and \method uses an earlier LoRA here.}

\label{fig:supp_cosmos_transfer}
\end{figure*}

\method trains one LoRA of rank 128, about 349M parameters, on the condition tokens of the released Cosmos3-Nano for the tasks in Tab.~\ref{tab:cosmos} and image-to-video.
It follows the standard OPSD setting of our multi-condition model, where the student rolls out the official sampler of each task, with 50 steps for transfer and 35 or 30 steps for the other tasks, and each training step supervises its first 8 states.
Each training step draws one task, with transfer on 40\% of the steps, continuation on 21\%, video-to-audio on 19\%, video-to-action on 15\%, and image-to-video on 5\%, and the clips vary in resolution and length.
We train for 3,000 steps with a batch size of 16 on 16 H200 GPUs and a learning rate of $10^{-4}$.
Figs.~\ref{fig:supp_cosmos_transfer} and~\ref{fig:supp_cosmos_av} are generated with an earlier LoRA.

\subsection{Qualitative results}
\label{sec:supp_cosmos_qual}

\begin{figure*}[p]
\centering
\includegraphics[width=\textwidth]{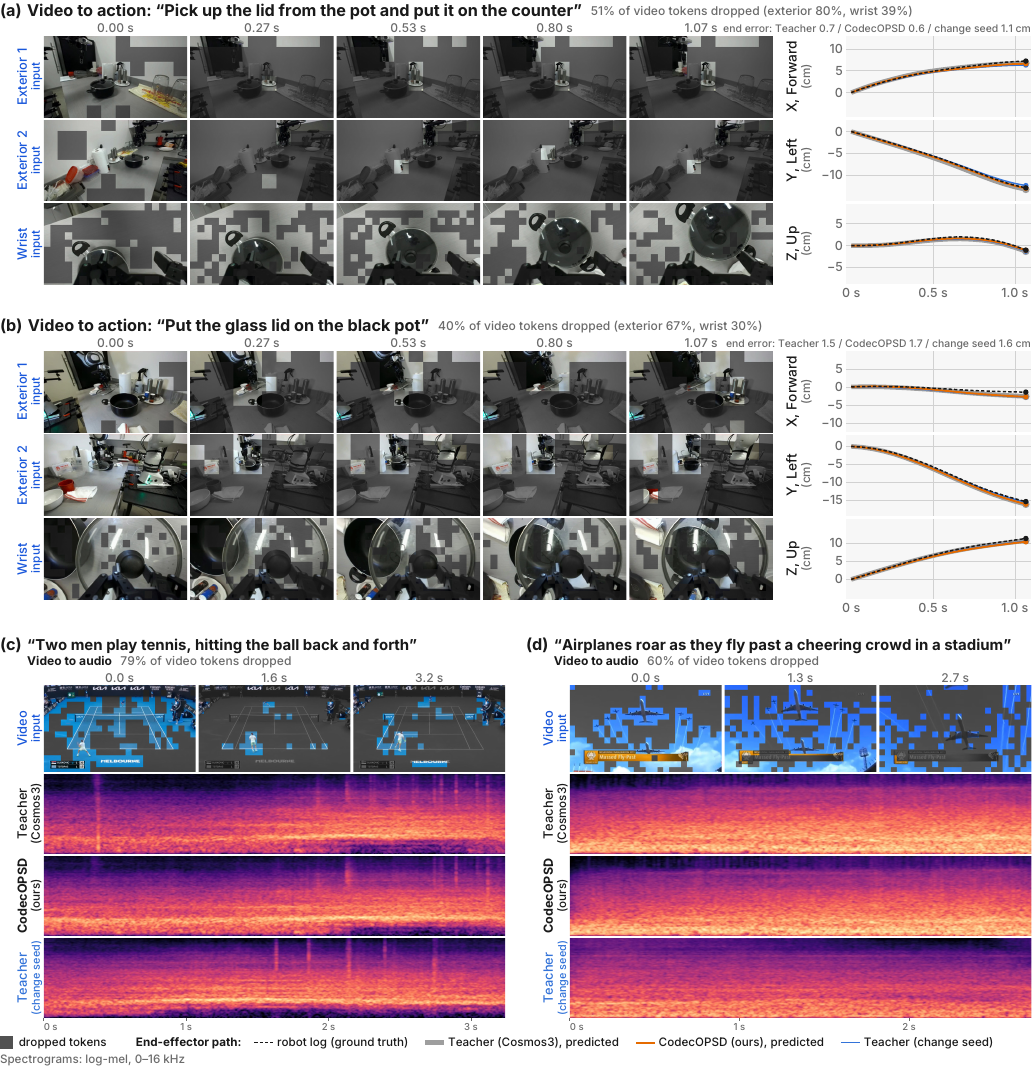}

\caption{\textbf{Codec pruning for video-to-action (robot inverse dynamics) and video-to-audio on Cosmos3-Nano~\cite{agarwal2026cosmos}.}
(a, b) DROID~\cite{khazatsky2024droid} episodes and (c, d) FoleyBench~\cite{dixit2026foleybench} clips show the input with the dropped video tokens darkened, followed by the end-effector path or the log-mel spectrogram of the teacher, of \method, and of the teacher with another noise seed.
\method stays close to the teacher in both tasks, and it uses an earlier LoRA than in Tab.~\ref{tab:cosmos}.}
\label{fig:supp_cosmos_av}

\end{figure*}

In Figs.~\ref{fig:supp_cosmos_transfer} and~\ref{fig:supp_cosmos_av}, we visualize the tokens dropped by codec pruning on Cosmos3-Nano.
Fig.~\ref{fig:supp_cosmos_transfer} combines all four transfer controls on robot clips, an extreme setting outside Tab.~\ref{tab:cosmos}.
Unlike the official pipeline of Cosmos-Transfer1~\cite{alhaija2025cosmos}, which runs one pass per control, we concatenate the tokens of all controls into one sequence.
This keeps one forward pass per step for an arbitrary number of controls, and codec pruning drops 80\% of the control tokens while \method stays close to the teacher.
In later frames, the static background is dropped, and the remaining tokens follow the moving arms.
For video-to-action, codec pruning drops 45\% of the input tokens, mostly from the static exterior views, and the action path of \method follows that of the teacher.
For video-to-audio, \method preserves the sound of the teacher with 60--79\% of the video tokens dropped.
A Qwen3-Omni judge considers the soundtracks of \method and the teacher to be the same sound with a probability of 0.79, compared with 0.20 for two runs of the teacher with different seeds.

\section{Limitations}
\label{sec:supp_limitations}

Overall, codec pruning works well across conditions and models, while we also observe a few limitations.
For motion tracks, codec pruning drops the tokens of a point that remains static after the frame where it first appears, so the remaining tokens may confuse a static point with an occluded or masked-out one.
In addition, a few detailed modalities, such as bounding boxes, motion tracks, and blurring, benefit from adjusted codec settings, as listed in Tab.~\ref{tab:supp_codec_settings}.
We believe that a learning-based search over these settings with pre-computation in advance for each modality is a promising direction for future work and may further improve the balance between speed and quality.

\end{document}